%% file: main.tex
\documentclass[10pt,twocolumn,letterpaper]{article}

 \usepackage[pagenumbers]{cvpr} 

\usepackage{stfloats}
\usepackage{graphicx}
\usepackage{amsmath}
\usepackage{amssymb}
\usepackage{bbding}
\usepackage{booktabs}
\usepackage{makecell}
\usepackage[export]{adjustbox}
\usepackage[normalem]{ulem}
\usepackage{textpos}  %
\usepackage[table]{xcolor}
\usepackage{accents}
\usepackage{longtable}

\usepackage{tabulary,multirow,overpic,xcolor,subfloat}

\definecolor{cvprblue}{rgb}{0.21,0.49,0.74}
\definecolor{linkcolor}{HTML}{ED1C24}
\usepackage[pagebackref,breaklinks,colorlinks,citecolor=cvprblue, linkcolor=linkcolor]{hyperref}

\newcommand{\app}{\raise.17ex\hbox{$\scriptstyle\sim$}}

\newcolumntype{x}[1]{>{\centering\arraybackslash}p{#1pt}}
\newcolumntype{y}[1]{>{\raggedright\arraybackslash}p{#1pt}}

\newlength\savewidth\newcommand\shline{\noalign{\global\savewidth\arrayrulewidth
		\global\arrayrulewidth 1pt}\hline\noalign{\global\arrayrulewidth\savewidth}}
\newcommand{\tablestyle}[2]{\setlength{\tabcolsep}{#1}\renewcommand{\arraystretch}{#2}\centering\footnotesize}

\newcommand{\figref}[1]{Fig.~\ref{#1}}
\newcommand{\secref}[1]{Sec.~\ref{#1}}
\newcommand{\tabref}[1]{Table~\ref{#1}}

\newcommand{\equref}[1]{Eq.~\ref{#1}}

\definecolor{Gray}{gray}{0.5}

\def\paperID{*****} 
\def\confName{CVPR}
\def\confYear{2024}

\title{\vskip -1.0em AnyStory: Towards Unified Single and Multiple Subject Personalization in Text-to-Image Generation\vspace{-0.5em}}

\vspace{-1em}
\author{Junjie He \ \ \ Yuxiang Tuo \ \ \ Binghui Chen \ \ \ Chongyang Zhong \ \ \ Yifeng Geng \ \ \ Liefeng Bo\\
	Institute for Intelligent Computing, Alibaba Tongyi Lab\\
	{\tt\small \{hejunjie.hjj, yuxiang.tyx\}@alibaba-inc.com \ \ \ chenbinghui@bupt.cn} \\
	{\tt\small \{zhongchongyang.zzy, cangyu.gyf, liefeng.bo\}@alibaba-inc.com}
}

\begin{document}
	
\makeatletter
\let\@oldmaketitle\@maketitle%
\renewcommand{\@maketitle}{\@oldmaketitle%
	\vskip -2.0em
	\centering
	\includegraphics[width=1\textwidth]{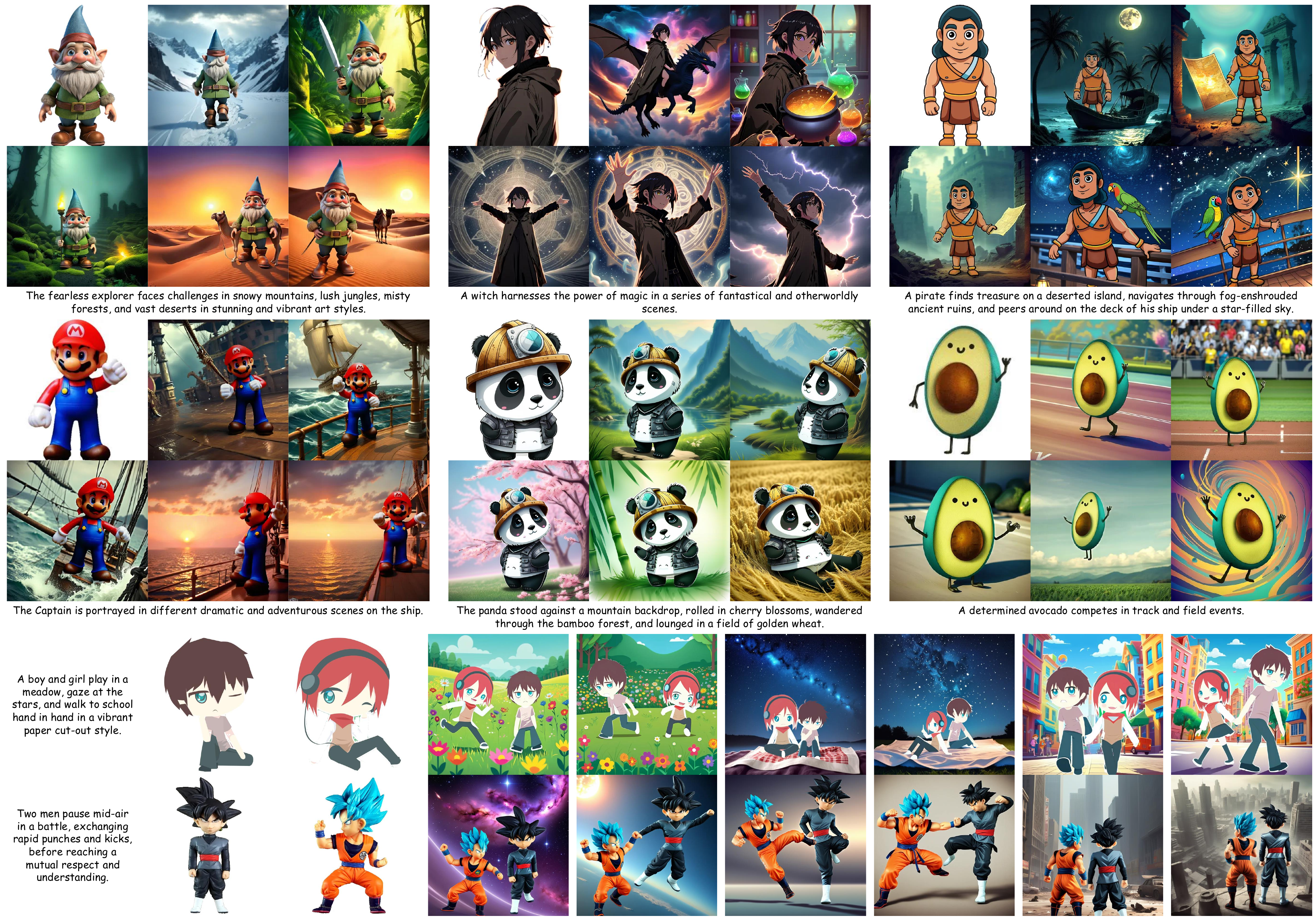}
	\vskip -0.5em
	\captionof{figure}{ 
		\textbf{Example generations \uppercase\expandafter{\romannumeral1} from AnyStory.}  Our approach demonstrates excellence in preserving subject details, aligning text descriptions, and personalizing multiple subjects. Here, the image with a plain white background serves as the reference. For more examples, please refer to \figref{fig:example-2} and \figref{fig:example-3}.
	}
	\label{fig:example-1}
	\bigskip}
\makeatother

\maketitle

\input{sections/abs}

\input{sections/intro}

\input{sections/relwork}
\input{sections/approach}

\input{sections/expts}

\input{sections/concl}

\clearpage
\clearpage
{
    \small
    \bibliographystyle{ieeenat_fullname}
    \bibliography{main}
}

\clearpage
\onecolumn
\appendix
\begin{center}{\bf \Large Appendix}\end{center}\vspace{-2mm}
\renewcommand{\thetable}{\Roman{table}}
\renewcommand{\thefigure}{\Roman{figure}}
\setcounter{table}{0}
\setcounter{figure}{0}

\input{sections/appendix}

\end{document}

%% file: sections/abs.tex
\begin{abstract}
	
Recently, large-scale generative models have demonstrated outstanding text-to-image generation capabilities. However, generating high-fidelity personalized images with specific subjects still presents challenges, especially in cases involving multiple subjects. In this paper, we propose AnyStory, a unified approach for personalized subject generation. AnyStory not only achieves high-fidelity personalization for single subjects, but also for multiple subjects, without sacrificing subject fidelity. Specifically, AnyStory models the subject personalization problem in an ``encode-then-route'' manner. In the encoding step, AnyStory utilizes a universal and powerful image encoder, \ie, ReferenceNet, in conjunction with CLIP vision encoder to achieve high-fidelity encoding of subject features. In the routing step, AnyStory utilizes a decoupled instance-aware subject router to accurately perceive and predict the potential location of the corresponding subject in the latent space, and guide the injection of subject conditions. Detailed experimental results demonstrate the excellent performance of our method in retaining subject details, aligning text descriptions, and personalizing for multiple subjects.
The project page is at \url{https://aigcdesigngroup.github.io/AnyStory/}.
\end{abstract}

%% file: sections/intro.tex
\section{Introduction}
\label{sec:intro}

Recently, with the rapid development of diffusion models~\cite{dhariwal2021diffusion,ho2020denoising,sohl2015deep,song2019generative}, many large generative models~\cite{podell2023sdxl, betker2023improving,ldm,li2024hunyuan,liu2024playground,chen2023pixart,chen2025pixart,peebles2023scalable} have demonstrated remarkable text-to-image generation capabilities. However, generating personalized images with specific subjects still presents challenges. Early efforts~\cite{avrahami2023break,chen2023disenbooth,textual_inversion,han2023svdiff,hu2022lora,kumari2023multi,ruiz2023dreambooth} utilize fine-tuning at test time to achieve personalized content generation. These methods require extensive fine-tuning time and their generalization ability is limited by the number and diversity of tuning images. Recent works~\cite{ye2023ip-adapter,li2024blipdiffusion,gal2023encoder,subject-diffusion,wei2023elite,shi2024instantbooth,zhang2024ssr,li2024photomaker,fastcomposer,wang2024instantid} have explored zero-shot settings. They have introduced specialized subject encoders to retrain text-to-image models on large-scale personalized image datasets, without the need for model fine-tuning at test time. However, these methods are either limited by the encoder's capability to provide high-fidelity subject details~\cite{ye2023ip-adapter,li2024blipdiffusion,subject-diffusion,wei2023elite,shi2024instantbooth,zhang2024ssr}, or focus on specific categories of objects (such as face identities~\cite{li2024photomaker,fastcomposer,wang2024instantid}) and cannot extend to general subjects (such as human clothing, accessories, and non-human entities), limiting their applicability.

In addition, previous methods mainly focus on single-subject personalization. Problems with subject blending often occur in multi-subject generation due to semantic leakage~\cite{fastcomposer,dahary2025yourself}. Some methods~\cite{liu2023cones,kim2024instantfamily,kwon2024concept,gu2024mix,wang2024instancediffusion,subject-diffusion,zhou2024migc,jang2024identity} address this issue by introducing pre-defined subject masks, but this restricts the diversity and creativity of generative models. Additionally, providing precise masks for subjects with complex interactions and occlusions is difficult. Recent research, \ie, UniPortrait~\cite{he2024uniportrait}, proposes a subject router to adaptively perceive and constrain the effect region of each subject condition in the diffusion denoising process. However, the routing features used by UniPortrait are highly coupled with subject identity features, limiting the accuracy and flexibility of the routing module. Furthermore, it primarily focuses on the domain of face identity and does not consider the impact of subject conditions on the background.

In this paper, we propose AnyStory, a unified single- and multi-subject personalization framework. We aim to personalize general subjects while achieving fine-grained control over multi-subject conditions. Additionally, we aim to allow the variation of subjects' backgrounds, poses, and views through text prompts while maintaining subject details, thus creating complex and fantastical narratives.

To achieve this, we have introduced two key modules, \ie, an enhanced subject representation encoder and a decoupled instance-aware subject router. 
To be specific, we adopt the ``encode-then-route'' design of UniPortrait.  In order to achieve a general subject representation, we abandon domain-specific expert models, such as the face encoders~\cite{deng2019arcface,huang2020curricularface}, and instead use a powerful and versatile model, \ie, ReferenceNet~\cite{hu2024animate}, combined with the CLIP vision encoder~\cite{radford2021learning} to encode the subject. CLIP vision encoder is responsible for encoding the subject's coarse concepts, while ReferenceNet is responsible for encoding the appearance details to enhance subject fidelity. To improve efficiency, we also simplify the architecture of ReferenceNet, skipping all cross-attention layers to save storage and computation costs. In order to avoid the copy-paste effect, we further collect a large amount of paired subject data, which is sourced from image, video, and 3D rendering databases. These paired data contain instances of the same subject in different contexts, effectively aiding the encoder in understanding and encoding provided subject concepts.

For the subject router, in contrast to UniPortrait, we implement a separate branch to allow for a specialized and flexible routing guidance. Additionally, we improve the structure of the routing module by modeling it as a mini-image segmentation decoder, with a masked cross-attention~\cite{cheng2022masked,he2023fastinst} and a background routing representation being introduced. Combined with instance-aware routing regularization loss, the proposed router can accurately perceive and predict the potential location of the corresponding subject in the latent during the denoising process. In practice, we observe that the behavior of this enhanced subject router is similar to image instance segmentation, which may provide a potential approach for image-prompted visual subject segmentation.

The experimental results demonstrate the outstanding performance of our method in preserving the fidelity of the subject details, aligning text descriptions, and personalizing for multiple subjects. Our contributions can be summarized as follows:

\begin{itemize}
	\item We propose a unified single- and multi-subject personalization framework called AnyStory. It achieves consistency in personalizing both single-subject and multi-subject while adhering to text prompts;

	\item We introduce an enhanced subject representation encoder, composed of a simplified lightweight ReferenceNet and CLIP vision encoder, capable of high-fidelity detail encoding for general subjects.

	\item We propose a decoupled instance-aware routing module that can accurately perceive and predict the potential conditioning areas of the subject, thereby achieving flexible and controllable personalized generation of single or multiple subjects.
\end{itemize}

%% file: sections/relwork.tex
\section{Related Work}

\textbf{Single-subject personalization.}
Personalized image generation with specific subjects is a popular and challenging topic in text-to-image generation. Early works~\cite{avrahami2023break,chen2023disenbooth,textual_inversion,gal2023designing,han2023svdiff,hu2022lora,kumari2023multi,ruiz2023dreambooth,voynov2023p+} rely on fine-tuning during testing. These methods typically require several minutes to even hours to achieve satisfactory results, and their generalization abilities are limited by the number of fine-tuned images. 
Recently, some methods~\cite{gal2023encoder,jia2023taming,li2024blipdiffusion,subject-diffusion,shi2024instantbooth,wei2023elite,ye2023ip-adapter,tan2024ominicontrol} have sought to achieve personalized image generation for subjects without additional fine-tuning.
IP-Adapter~\cite{ye2023ip-adapter} encodes subjects into text-compatible image prompts for subject personalization.
BLIP-Diffusion~\cite{li2024blipdiffusion} introduces a pre-trained multimodal encoder to provide subject representation.
SSR-Encoder~\cite{zhang2024ssr} proposes a token-to-patch aligner and detail-preserved subject encoder to learn selective subject embedding.
FaceStudio~\cite{yan2023facestudio}, InstantID~\cite{wang2024instantid}, and PhotoMaker~\cite{li2024photomaker} utilize face embeddings derived from face encoders as the condition.
Although these methods have made progress, they are either limited by the ability of the image encoder to preserve subject details~\cite{ye2023ip-adapter,li2024blipdiffusion,gal2023encoder,subject-diffusion,wei2023elite,shi2024instantbooth,zhang2024ssr}, or focus on specific domains, \eg, face identity, without the ability to generalize to other objects~\cite{li2024photomaker, wang2024instantid, yan2023facestudio, he2024uniportrait,guo2024pulid}.

\noindent\textbf{Multi-subject personalization.} 
Significant progress has been made in single-subject personalization. However, the personalized generation of multi-subject images still presents challenges due to the problem of subject blending~\cite{fastcomposer,dahary2025yourself}. To overcome these challenges, recent studies~\cite{liu2023cones,kim2024instantfamily,kwon2024concept,gu2024mix,wang2024instancediffusion,subject-diffusion,zhou2024migc} have utilized predefined layout masks to guide multi-subject generation. 
However, these layout-dependent methods limit the creativity of the generation models and the diversity of resulting images. Additionally, providing precise layout masks for each subject in complex contexts is challenging. 
Some methods obtain subject masks from attention maps corresponding to subject tokens~\cite{shen2024rethinking,wang2023diffusion,chefer2023attend,hertz2022prompt,tewel2024training,cao2023masactrl} or from segmentation of existing images~\cite{kong2024omg,he2024dreamstory}, which may result in inaccurate masks for the target subject instances.
FastComposer~\cite{fastcomposer}, Subject-Diffusion~\cite{subject-diffusion}, and StoryMaker~\cite{zhou2024storymaker} impose constraints on cross-attention maps for different subjects during training, but this impacts the injection of subject conditions. Recently, UniPortrait~\cite{he2024uniportrait} introduces a subject router to perceive and predict subject potential positions during denoising, avoiding blending adaptively. However, its routing features are highly coupled with subject features, limiting the precision of the routing module. 

\noindent\textbf{Story visualization.} 
Generating visual narratives based on given scripts, known as story visualization~\cite{avrahami2024chosen,tewel2024training,he2024dreamstory,zhou2024storymaker}, is rapidly evolving. 
StoryDiffusion~\cite{zhou2024storydiffusion} proposes a consistent self-attention calculation to ensure the consistency of characters throughout the story sequence.
ConsiStory~\cite{tewel2024training} proposes a training-free approach that shares the internal activations of the pre-trained diffusion model to achieve subject consistency. 
DreamStory~\cite{he2024dreamstory} utilizes a Large Language Model (LLM) and a multi-subject consistent diffusion model, incorporating masked mutual self-attention and masked mutual cross-attention modules,  to generate consistent multi-subject story scenes.
The proposed method in this paper achieves subject consistency in image sequence generation through routed subject conditioning.

%% file: sections/approach.tex
\begin{figure*}[t]
	\centering
	\includegraphics[width=1.0\linewidth]{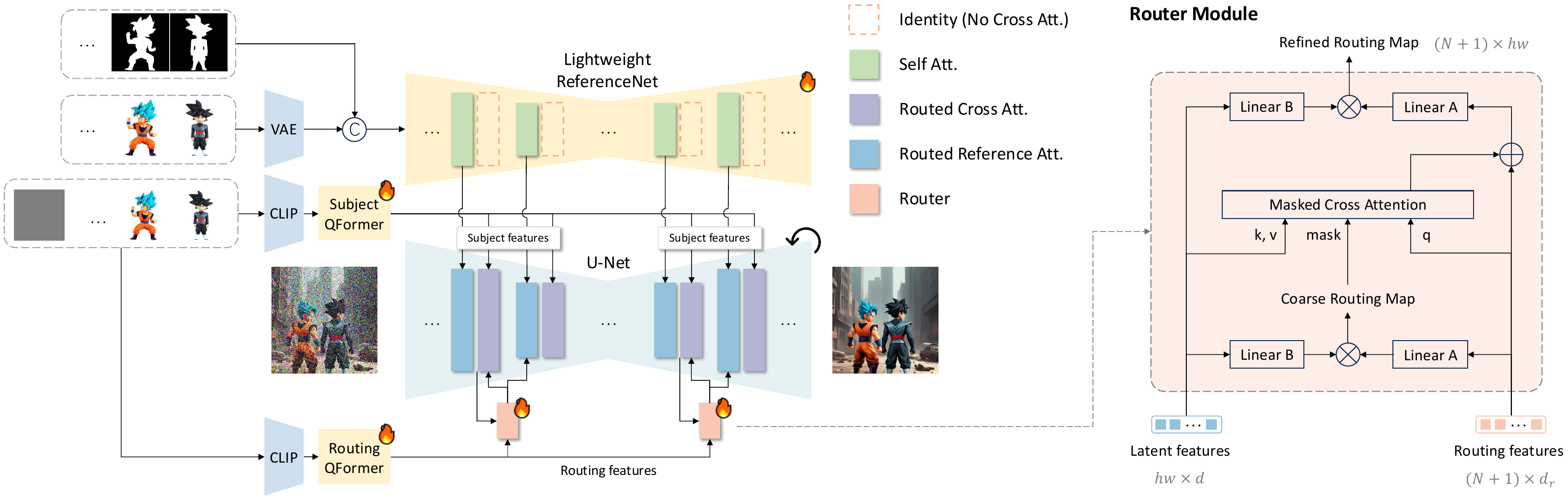}
	\caption{
		\textbf{Overview of AnyStory framework}. 
		AnyStory follows the ``encode-then-route'' conditional generation paradigm. It first utilizes a simplified ReferenceNet combined with a CLIP vision encoder to encode the subject (\secref{sec:3.2}), and then employs a decoupled instance-aware subject router to guide the subject condition injection (\secref{sec:3.3}). The training process is divided into two stages: the subject encoder training stage and the router training stage (\secref{sec:3.4}). For brevity, we omit the text conditional branch here.
	}
	\label{fig:framework}
\end{figure*}

\section{Methods}

We introduce AnyStory, a pioneering method for unified single- and multi-subject personalization in text-to-image generation. 
We first briefly review the background of the diffusion model in \secref{sec:3.1}, and then detail the two proposed key components, \ie, the enhanced subject encoder and the decoupled instance-aware subject router, in \secref{sec:3.2} and \secref{sec:3.3}, respectively. Finally, we outline our training scheme in \secref{sec:3.4}.  The framework of our method is illustrated in \figref{fig:framework}.

\subsection{Preliminary}
\label{sec:3.1}

The underlying text-to-image model we used in this paper is Stable Diffusion XL (SDXL)~\cite{podell2023sdxl}. SDXL takes a text prompt $P$ as input and produces the image $x_0$. It contains three modules: an autoencoder $(\mathcal{E}(\cdot), \mathcal{D}(\cdot))$, a CLIP text encoder $\tau(\cdot)$, and a U-Net $\epsilon_{\theta}(\cdot)$. Typically, it is trained using the following diffusion loss:
\begin{equation}
	\mathcal{L}_{diff} = \mathbb{E}_{z_0,P,\epsilon \sim \mathcal{N}(0,1),t}\lbrack\Vert\epsilon-\epsilon_\theta (z_t,t,\tau(P))\Vert_2^2\rbrack
	\label{eq:1}
\end{equation}
where $\epsilon \sim \mathcal{N}(0,1)$ is the sampled Gaussian noise, $t$ is the time step,  $z_0=\mathcal{E}(x_0)$ is the latent code of $x_0$, and $z_t$ is computed by $z_t=\alpha_t z_0+\sigma_t\epsilon$ with the coefficients $\alpha_t$ and $\sigma_t$ provided by the noise scheduler.

\subsection{Enhanced subject representation encoding}
\label{sec:3.2}

Personalizing subject images in an open domain while ensuring fidelity to subject details and textual descriptions remains an unresolved issue. A key challenge lies in the encoding of subject information, which requires maximal preservation of subject characteristics while maintaining a certain level of editing capability. Current mainstream methods~\cite{ye2023ip-adapter,li2024blipdiffusion,gal2023encoder,subject-diffusion,wei2023elite,shi2024instantbooth,zhang2024ssr,liu2023cones} largely rely on CLIP vision encoder to encode subjects. 
However, CLIP's features are primarily semantic (for the reason of contrastive image-text training paradigm) and of  low-resolution (typically $224\times 224$), thus limited to providing thorough details of the subjects.
Alternative approaches~\cite{li2024photomaker,peng2024portraitbooth,guo2024pulid,wang2024instantid} incorporate domain-specific expert models, such as face encoders~\cite{deng2019arcface,huang2020curricularface}, to enhance subject identity representation. Despite their success, they are limited in their domain and are not extendable to general subjects.
To address these issues, we introduce ReferenceNet~\cite{hu2024animate}, a powerful and versatile image encoder, to encode the subject in conjunction with the CLIP vision encoder. ReferenceNet utilizes a variational autoencoder (VAE)~\cite{kingma2013auto,podell2023sdxl} to encode reference images and then extracts their features through a network with the same architecture as U-Net. It boasts three prominent advantages: (1) it supports higher resolution inputs, thereby enabling it to retain more subject details; (2) it has a feature space aligned with the denoising U-Net, facilitating the direct extraction of subject features at different depths and scales by U-Net; (3) it uses pre-trained U-Net weights for initialization, 
which possess a wealth of visual priors and demonstrate good generalization ability for learning general subject concepts.

\noindent\textbf{CLIP encoding.}
Following previous approaches~\cite{ye2023ip-adapter,he2024uniportrait}, we utilize the hidden states from the penultimate layer of the CLIP image encoder, which align well with image captions, as a rough visual concept representation of the subject. We first segment the subject area in the reference image to remove background information, and then input the segmented image into the CLIP image encoder to obtain a 257-length patch-level feature $\mathbf{F}_{\mathrm{clip}}$. Subsequently, we compress $\mathbf{F}_{\mathrm{clip}}$ using a QFormer~\cite{li2023blip,alayrac2022flamingo} into $m$ tokens. The final result, denoted as $\mathbf{E}$, serves as the subject representation derived from the CLIP vision encoder:
\begin{equation}
	\mathbf{E}=\operatorname{QFormer}(\mathbf{F}_\mathrm{clip}),
	\label{eq:2}
\end{equation}
where $\mathbf{E}\in \mathbb{R} ^{m\times d_c}$, and $d_c$ is the same as the text feature dimension in the pre-trained diffusion model. Empirically, we set $m$ to 64 in our experiments.

\noindent\textbf{ReferenceNet encoding.}
In the original implementation~\cite{hu2024animate}, ReferenceNet adopts the same architecture as U-Net, including cross-attention blocks with text condition injection. However, since ReferenceNet is only used as a visual feature extractor in our task and does not require text condition injection, we skip all cross-attention blocks, reducing the number of parameters and computational complexity (see \tabref{tab:referencenet}). Additionally, in order to label the subject areas, we also add a subject mask channel to the input of ReferenceNet. 
Specifically, we feed the segmented subject reference to the VAE encoder for encoding and then concatenate the encoded result with the downsampled subject mask to obtain $\mathbf{F}_\mathrm{vae}$. 
Next, $\mathbf{F}_\mathrm{vae}$ undergoes the reduced non-cross attention ReferenceNet.
The hidden states of each self-attention layer, denoted as $\{\mathbf{G}^{l} | \mathbf{G}^{l}\in\mathbb{R}^{h_Gw_G\times d}\}_l$, are extracted as the ReferenceNet-encoded representation of the subject:
\begin{equation}
	\{\mathbf{G}^{l}\}_{l=2}^L=\operatorname{ReferenceNet}(\mathbf{F}_\mathrm{vae}),
	\label{eq:3}
\end{equation}
where $l$ denotes the layer index, and $L$ denotes the total number of layers. Here, we ignore the features of the first self-attention layer in order to better align with the routing module (see \secref{sec:3.3}).

\begin{table}[t]
	\centering
	\tablestyle{4pt}{1.2}\scriptsize\begin{tabular}{x{86} | x{40} x{60}  }
		{Architecture} &  {\#Params (B)} & {Speed (ms/img)}   \\
		\shline
		Original ReferenceNet~\cite{hu2024animate} &  2.57 & 62.0  \\
		Simplified  ReferenceNet & 2.02 & 53.2 \\	
	\end{tabular}
	\caption{\textbf{Statistics of the simplified ReferenceNet.} The speed is measured on an A100 GPU with a batch size of 1 and an input (latent) resolution of $64\times64$.}
	\label{tab:referencenet}
\end{table}

\subsection{Decoupled instance-aware subject routing}
\label{sec:3.3}

The injection of subject conditions requires careful consideration of injecting positions to avoid the influence on unrelated targets.
Previous methods~\cite{ye2023ip-adapter,kumari2023multi,li2024blipdiffusion,zhang2024ssr,shi2024instantbooth,wei2023elite} have typically injected the conditional features into the latent through a naive attention module.
However, due to the soft weighting mechanism, these approaches are prone to semantic leakage~\cite{fastcomposer,dahary2025yourself}, leading to subject characteristics blending, especially in the generation of instances with similar appearance. 
Some methods~\cite{liu2023cones,kim2024instantfamily,kwon2024concept,gu2024mix,wang2024instancediffusion,subject-diffusion,zhou2024migc} introduce predefined layout masks to address this issue, but this limits their practical applicability. 
UniPortrait~\cite{he2024uniportrait} proposes a router to perceive and confine the effect region of subject conditions adaptively; however, its routing features are completely coupled with subject features, which limits the ability of the routing module; also,  it does not consider the impact of subject conditions on the background. In this study, we propose a decoupled instance-aware subject routing module, which can accurately and effectively route subject features to the corresponding areas while reducing the impact on unrelated areas.

\noindent\textbf{Decoupled routing mechanism.}
Different from UniPortrait~\cite{he2024uniportrait}, we employ an independent branch to specifically predict the potential location of subjects in the latent during the denoising process. 
As depicted in the \figref{fig:framework}, given a series of segmented subject images, they are respectively passed through CLIP image encoder and an additional one-query QFormer to obtain the routing features $\{\mathbf{R}_i |\mathbf{R}_i\in\mathbb{R}^{d_r}\}_{i=1}^{N}$, where $N$ represents the number of reference subjects.
Particularly, we include an additional routing feature $\mathbf{R}_{N+1}$ for the background (zero image as input) to further confine the subject's conditioning areas. 
The idea behind this is to alleviate the undesirable biases inherent in subject features on the generated image backgrounds (e.g., we used a large amount of pure white background data from 3D rendering to train the subject encoder).
To accurately route the subjects to their respective positions, we employ an image segmentation decoder~\cite{cheng2022masked, he2023fastinst} to model the router. Specifically, in each cross-attention layer of the  U-Net, we first predict a coarse routing map by taking the linearly projected inner product of 
$\{\mathbf{R}_i \}_{i=1}^{N+1}$ and $\mathbf{Z}^l$. Here, $\mathbf{Z}^l\in\mathbb{R}^{hw\times d}$ represents the latent features in the $l$-th layer.
Subsequently, we refine the routing features $\{\mathbf{R}_i \}_{i=1}^{N+1}$
using a masked cross attention~\cite{cheng2022masked} with the latent feature $\mathbf{Z}^l$, where the coarse routing map serves as the attention mask. The updated routing features are then subjected to the projected inner product with $\mathbf{Z}^l$ again to obtain the refined routing maps
$\{\mathbf{M}_i^l \ |\ \mathbf{M}_i^l \in [0, 1]^{hw}\}_{i=1}^{N+1}$. $\{\mathbf{M}_i^l\}_{i=1}^{N+1}$
are finally used to guide the injection of information related to the subjects at that layer. For a detailed structure of the router, please refer to the right half of \figref{fig:framework}.

\noindent\textbf{Instance-aware routing regularization loss.}
In order to facilitate router learning and to differentiate between different instances of the subjects, we introduce an instance-aware routing regularization loss. The loss function is defined as:
\begin{equation}
	L^{l}_{route} = \lambda \cdot \frac{1}{N} \sum_{i=1}^{N} || \mathbf{M}_i^l - \mathbf{M}_i^{gt}||_2^2
	\label{eq:4}
\end{equation}
where $\mathbf{M}_i^{gt}\in\{0,1\}^{hw}$ represents the downsampled ground truth  mask of the $i$-th subject in the target image. Typically, we consider the entire subject instance, such as the full human body, as the routing target, regardless of whether the input subject has been cropped. 

\noindent\textbf{Routing-guided subject information injection.} For CLIP encoded subject representations, we use the decoupled cross attention~\cite{ye2023ip-adapter} to incorporate them into the U-Net, but with additional routing-guided localization constraints:
\begin{equation}
	\begin{aligned}
	\hat{\mathbf{Z}}^l=&\operatorname{Softmax}(\frac{\mathbf{Q} \mathbf{K}^T}{\sqrt{d}})\mathbf{V} \\ &+ \eta\sum_{i=1}^{N+1}{\sigma(\mathbf{M}_i^l) \odot \operatorname{Softmax}(\frac{\mathbf{Q} {\hat{\mathbf{K}}^l_i}^T}{\sqrt{d}})\hat{\mathbf{V}}_i^l},
\end{aligned}
\label{eq:5}
\end{equation}
where $\mathbf{Q}=\mathbf{Z}^l\mathbf{W}_q^l$, $\mathbf{K}=\mathbf{C}\mathbf{W}_k^l$, and $\mathbf{V}=\mathbf{C}\mathbf{W}_v^l$
represent the query, key, and value matrices for text conditions, $\mathbf{C}$ represents text embeddings, $\hat{\mathbf{K}}_i^l=\mathbf{E}_i \hat{\mathbf{W}}_k^l$ and $\hat{\mathbf{V}}_i^l={\mathbf{E}}_i \hat{\mathbf{W}}_v^l$ represent the key and value matrices for the CLIP-encoded $i$-th subject, $\hat{{\mathbf{W}}}_k^l$ and $\hat{{\mathbf{W}}}_v^l$ are both trainable parameters, $\sigma(\mathbf{M}_i^l)$ represents the ``0-1'' version of $\mathbf{M}_i^l$ after operations \texttt{argmax} and \texttt{one-hot} for $\{\mathbf{M}_i^l\}_i$ over $i$ dimension, $\odot$ represents element-wise multiplication, and $\eta$ represents the strength of conditions.
It should be noted that here we also include an additional background representation, \ie,
$\mathbf{E}_{N+1}$, from CLIP, which similarly corresponds to zero-valued image inputs. This embedding ($\mathbf{E}_{N+1}$) is also utilized as an unconditional embedding during classifier-free guidance sampling training~\cite{ho2022classifier}.

In regard to the injection of ReferenceNet encoded subject features, we adopt the original reference attention~\cite{hu2024animate} but with an additional attention mask induced from routing maps. With a slight abuse of notation, this process can be formulated as follows:
\begin{equation}
	\begin{aligned}
	\tilde{\mathbf{Z}}^l=&\operatorname{Softmax}(\frac{\mathbf{Q} {[\mathbf{K}, \tilde{\mathbf{K}}_1^l,\cdots,\tilde{\mathbf{K}}_N^l]}^T}{\sqrt{d}} \\ &+ \mathrm{Bias}(\{\mathbf{M}_i^{l-1}\}_i,\gamma))[\mathbf{V}, \tilde{\mathbf{V}}_1^l,\cdots,\tilde{\mathbf{V}}_N^l],
\end{aligned}
\label{eq:6}
\end{equation}
where $\mathbf{Q}=\mathbf{Z}^l\mathbf{W}_q^l$, $\mathbf{K}=\mathbf{Z}^l\mathbf{W}_k^l$,  and $\mathbf{V}=\mathbf{Z}^l\mathbf{W}_v^l$ represent the query, key, and value matrices for self-attention, $\tilde{\mathbf{K}}_i^l=\mathbf{G}_i^l \tilde{\mathbf{W}}_k^l$ and $\tilde{\mathbf{V}}_i^l={\mathbf{G}}_i^l \tilde{\mathbf{W}}_v^l$ indicate the key and value matrices for the ReferenceNet-encoded features of the $i$-th subject at the $l$-th layer, $[\cdot]$ represents \texttt{concat} operation, $\mathrm{Bias}(\{\mathbf{M}_i^{l-1}\}_i,\gamma)$ represents the applied attention bias,
\begin{equation}
	\mathrm{Bias}(\{\mathbf{M}_i^{l-1}\}_i,\gamma)=[\mathbf{0},g(\mathbf{M}_1^{l-1})+\gamma,\cdots,g(\mathbf{M}_N^{l-1})+\gamma],
	\label{eq:7}
\end{equation}
where $\gamma$ controls the overall strength of ReferenceNet conditions,
$g(\mathbf{M}_i^{l-1})\in\{0,-\infty\}^{hw\times h_Gw_G}$
represents the attention bias derived from the routing maps of the preceding cross-attention layer, and its specific calculation is as follows:
\begin{equation}
	g(\mathbf{M}_i^{l-1})_{u,v}=\left\{
	\begin{array}{cl}
		0 &  \mathrm{if}\;\sigma(\mathbf{M}_i^{l-1})_u = 1 \\
		-\infty  &  \mathrm{otherwise} \\
	\end{array}. \right.
	\label{eq:8}
\end{equation}
Similar to UniPortrait, to ensure proper gradient backpropagation through $\sigma(\cdot)$ during training, we employ the Gumbel-softmax trick~\cite{jang2016categorical}.
In practice, we observed that the routing map behaves similarly to the instance segmentation mask, providing a potential method for reference-prompted image segmentation (first encode the image with VAE, then feed the encoded image and reference into denoising U-Net and router respectively to predict the masks, see \figref{fig:router-visualize}).

\begin{figure}[t]
	\centering
	\includegraphics[width=1.0\linewidth]{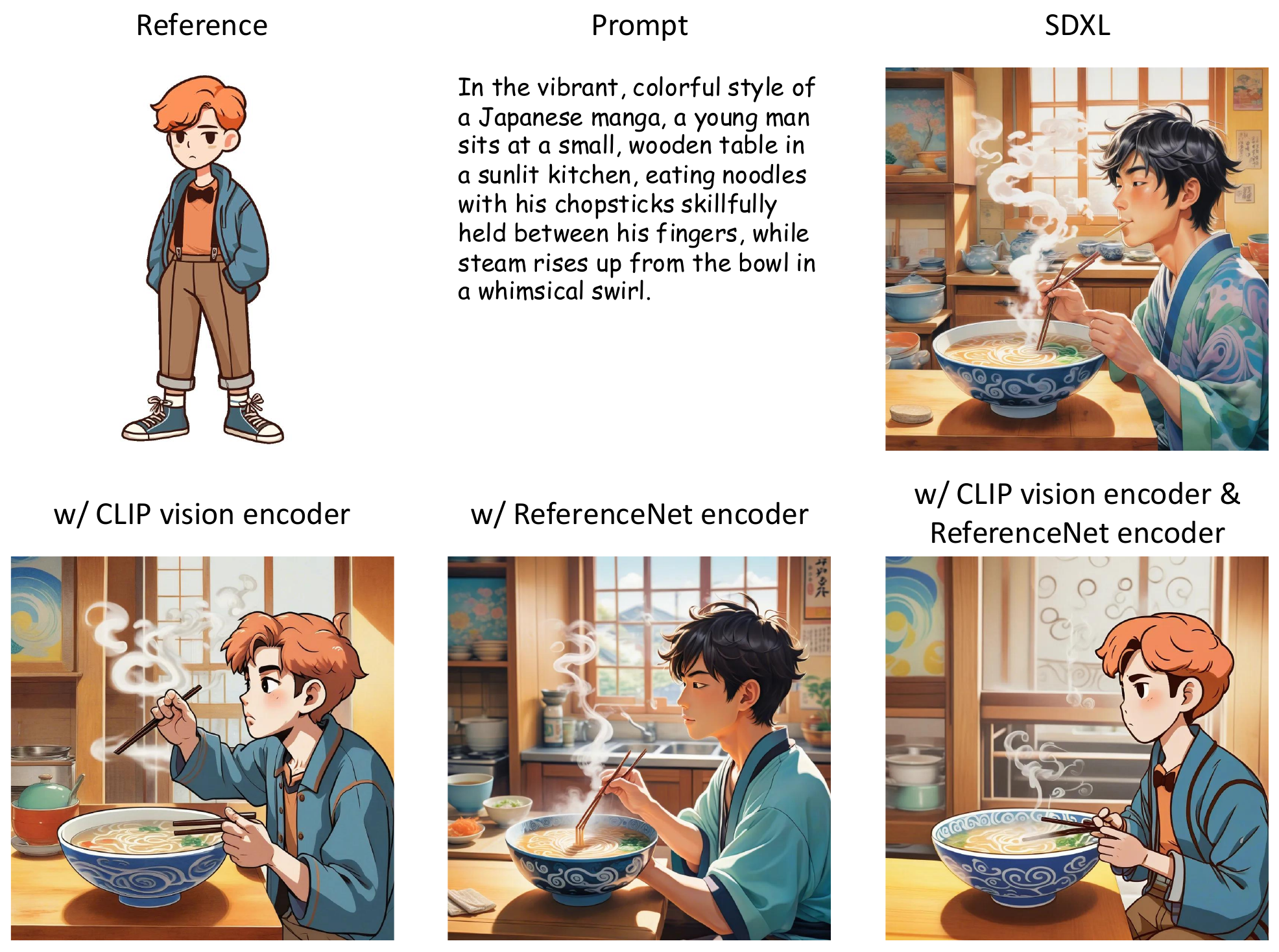}
	\caption{
		\textbf{Effect of ReferenceNet encoding.}  The ReferenceNet encoder enhances the preservation of subject details.
	}
	\label{fig:referencenet-encoder}
\end{figure}

\subsection{Training}
\label{sec:3.4}

Following UniPortrait, the training process of AnyStory is divided into two stages:  subject encoder training stage and router training stage.

\noindent\textbf{Subject encoder training.} We train the subject QFormer, ReferenceNet, and corresponding key, value matrices in attention blocks. The ReferenceNet utilizes pre-trained U-Net weights for initialization. 
To avoid the copy-paste effect caused by fine-grained encoding of subject features, we collect a large amount of paired data that maintains consistent subject identity while displaying variations in background, pose, and views. These data are sourced from image, video, and 3D rendering databases, captioned by
Qwen2-VL~\cite{wang2024qwen2}. Specifically, the image (about 410k) and video (about 520k) data primarily originate from human-centric datasets such as DeepFashion2~\cite{ge2019deepfashion2} and human dancing videos, while the 3D data (about 5,600k) is obtained from the Objaverse~\cite{deitke2023objaverse}, where images of objects from seven different perspectives are rendered as paired data. During the training process, one image from these pairs is utilized as the reference input, while another image, depicting the same subject identity but in a different context, serves as the prediction target. Additionally, data augmentation techniques, including random rotation, cropping, and zero-padding, are applied to the reference image to further prevent subject overfitting.  The training loss in this stage is the same as the original diffusion loss, as shown in \equref{eq:1}.

\noindent\textbf{Router training.}
We fix the subject encoder and train the router. The primary training data consists of an additional 300k unpaired multi-person images from LAION~\cite{schuhmann2022laion,schuhmann2021laion}. Surprisingly, despite the training dataset of the router being predominantly focused on human images, it is able to effectively generalize to general subjects. We attribute this to the powerful generalization ability of the CLIP model and the highly compressed single-token routing features. The training loss for this stage includes the diffusion loss (\equref{eq:1}) and the routing regularization loss (\equref{eq:4}), with the balancing parameter $\lambda$ set to 0.1.

%% file: sections/expts.tex
\begin{figure}[t]
	\centering
	\includegraphics[width=1.0\linewidth]{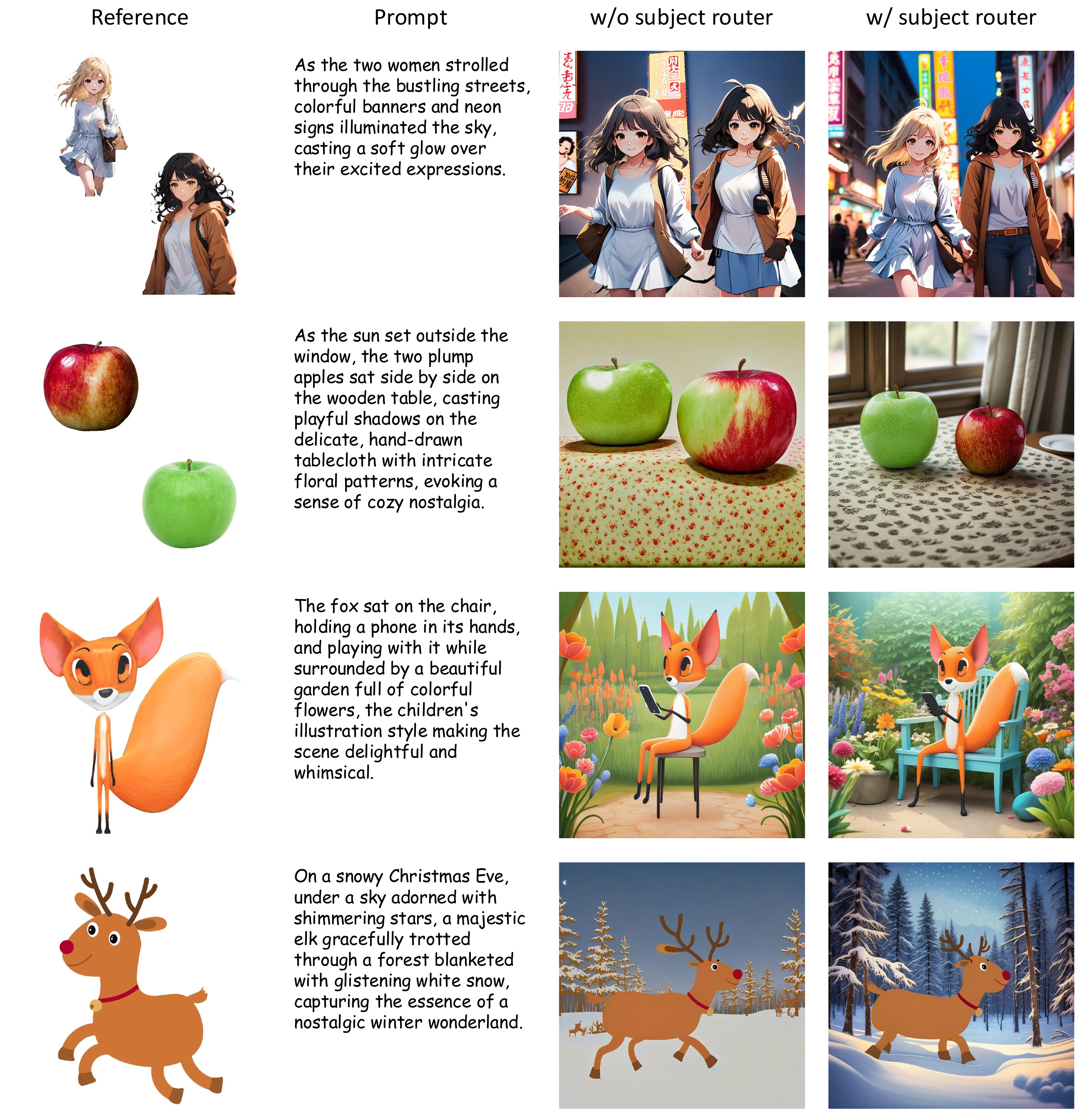}
	\caption{
		\textbf{The effectiveness of the router.} The router restricts the influence areas of the subject conditions, thereby avoiding the blending of characteristics between multiple subjects and improving the quality of the generated images.
	}
	\label{fig:router-effect}
\end{figure}

\begin{figure*}[t]
	\centering
	\includegraphics[width=1.0\linewidth]{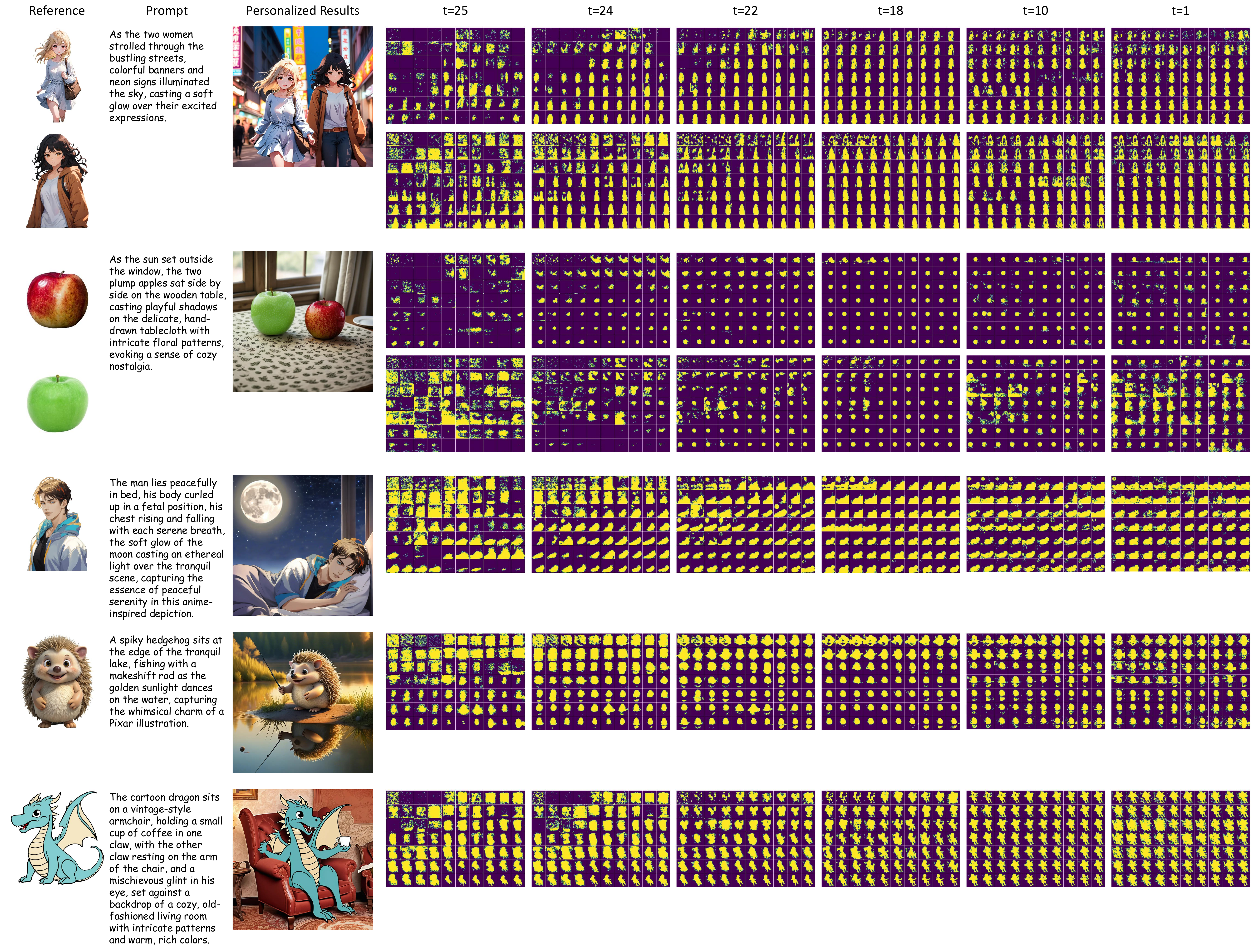}
	\caption{
		\textbf{Visualization of routing maps.}  
		We visualize the routing maps within each cross-attention layer in the U-Net at different diffusion time steps. There are a total of 70 cross-attention layers in the SDXL U-Net, and we sequentially display them in each subfigure in a top-to-bottom and left-to-right order (yellow represents the effective region). We utilize $T=25$ steps of EDM sampling. Each complete row corresponds to one entity. The background routing map has been ignored, which is the complement of the routing maps of all subjects. Best viewed in color and zoomed in.
	}
	\label{fig:router-visualize}
\end{figure*}

\begin{figure*}[t]
	\begin{subtable}{1.0\linewidth}
		\centering
		\bgroup
		\def\arraystretch{0.2}
		\setlength\tabcolsep{0.2pt}
		\begin{tabular}{c}
			\begin{subtable}{1.0\linewidth}
				\includegraphics[width=1.0\linewidth]{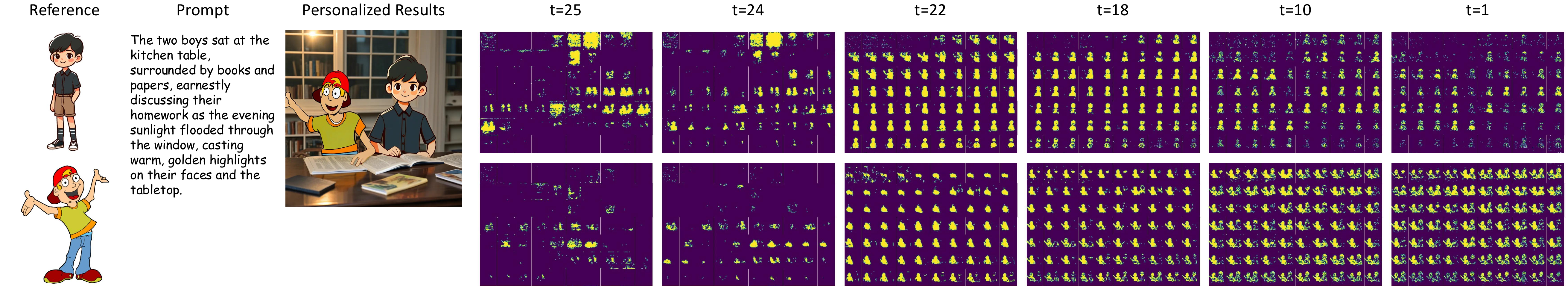} 
				\caption{Coarse routing maps}
				\label{fig:router-coarse}
			\end{subtable}
			\vspace{0.5em}
			
			\\
			\begin{subtable}{1.0\linewidth}
				\includegraphics[width=1.0\linewidth]{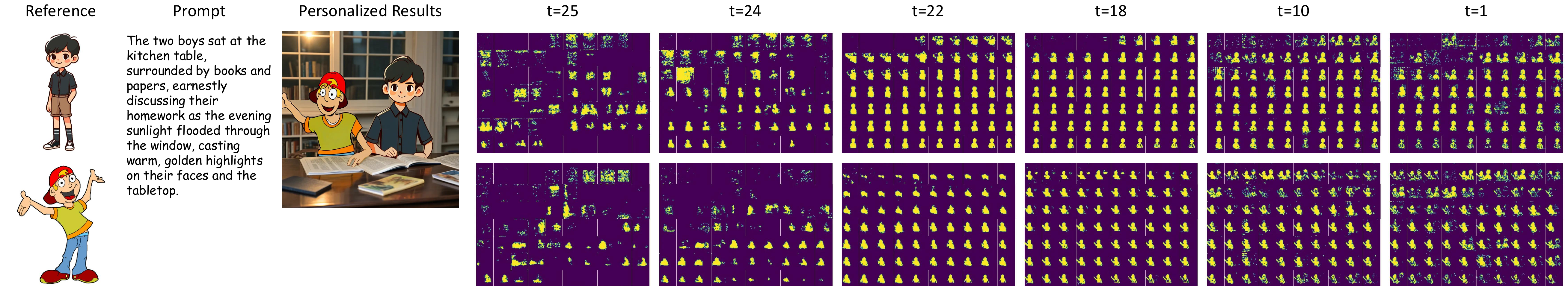} 
				\caption{Refined routing maps}
				\label{fig:router-refine}
			\end{subtable}
		\end{tabular} \egroup
	\end{subtable}
	\caption{\textbf{Effectiveness of the proposed router structure.}  For the meaning of each illustration, please refer to  \figref{fig:router-visualize}.}
	\label{fig:router-coarse-refine}
\end{figure*}

\begin{figure*}[t]
	\centering
	\includegraphics[width=1.0\linewidth]{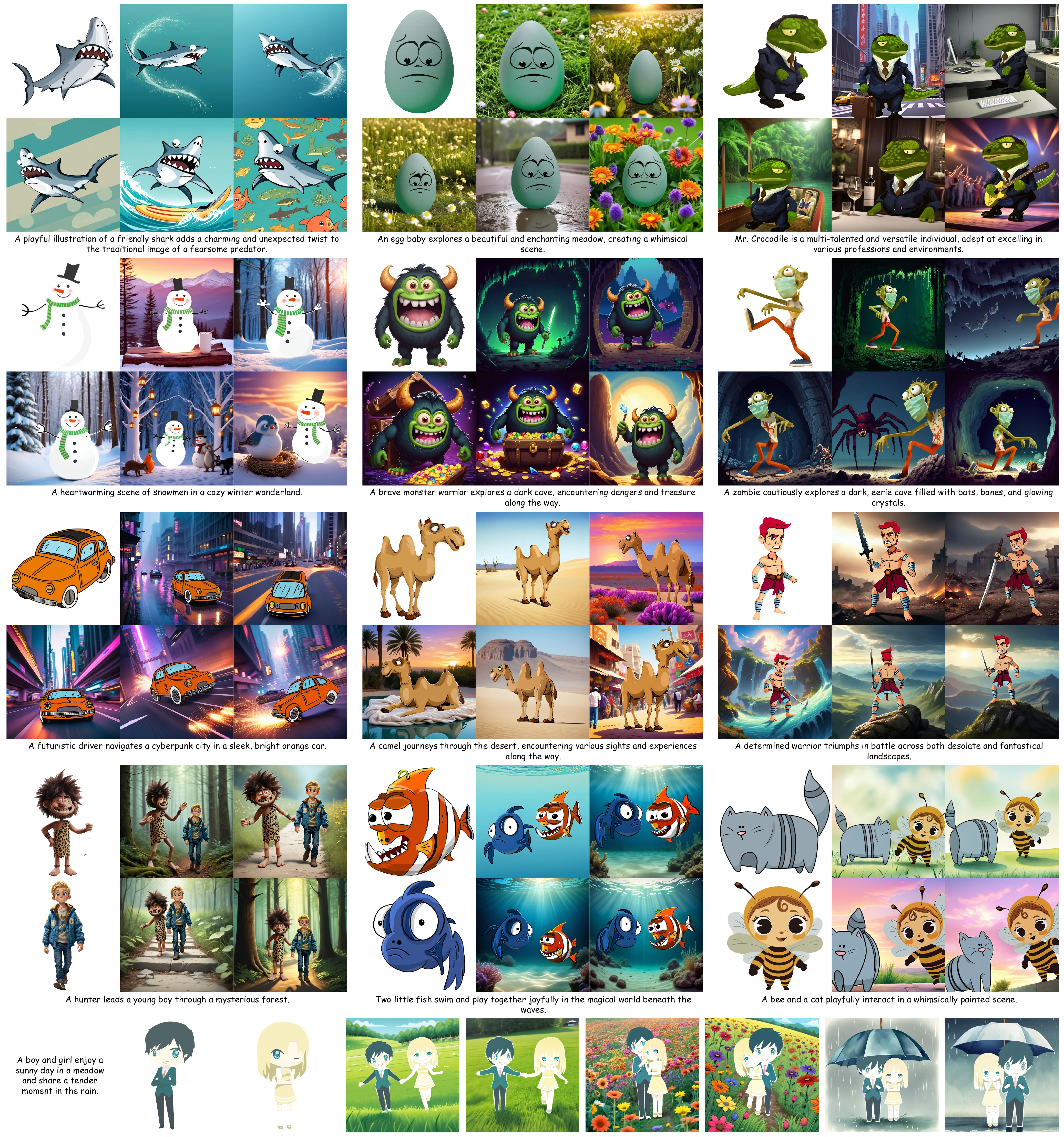}
	\caption{
		\textbf{Example generations \uppercase\expandafter{\romannumeral2} from AnyStory.} 
	}
	\label{fig:example-2}
\end{figure*}

\begin{figure*}[t]
	\centering
	\includegraphics[width=1.0\linewidth]{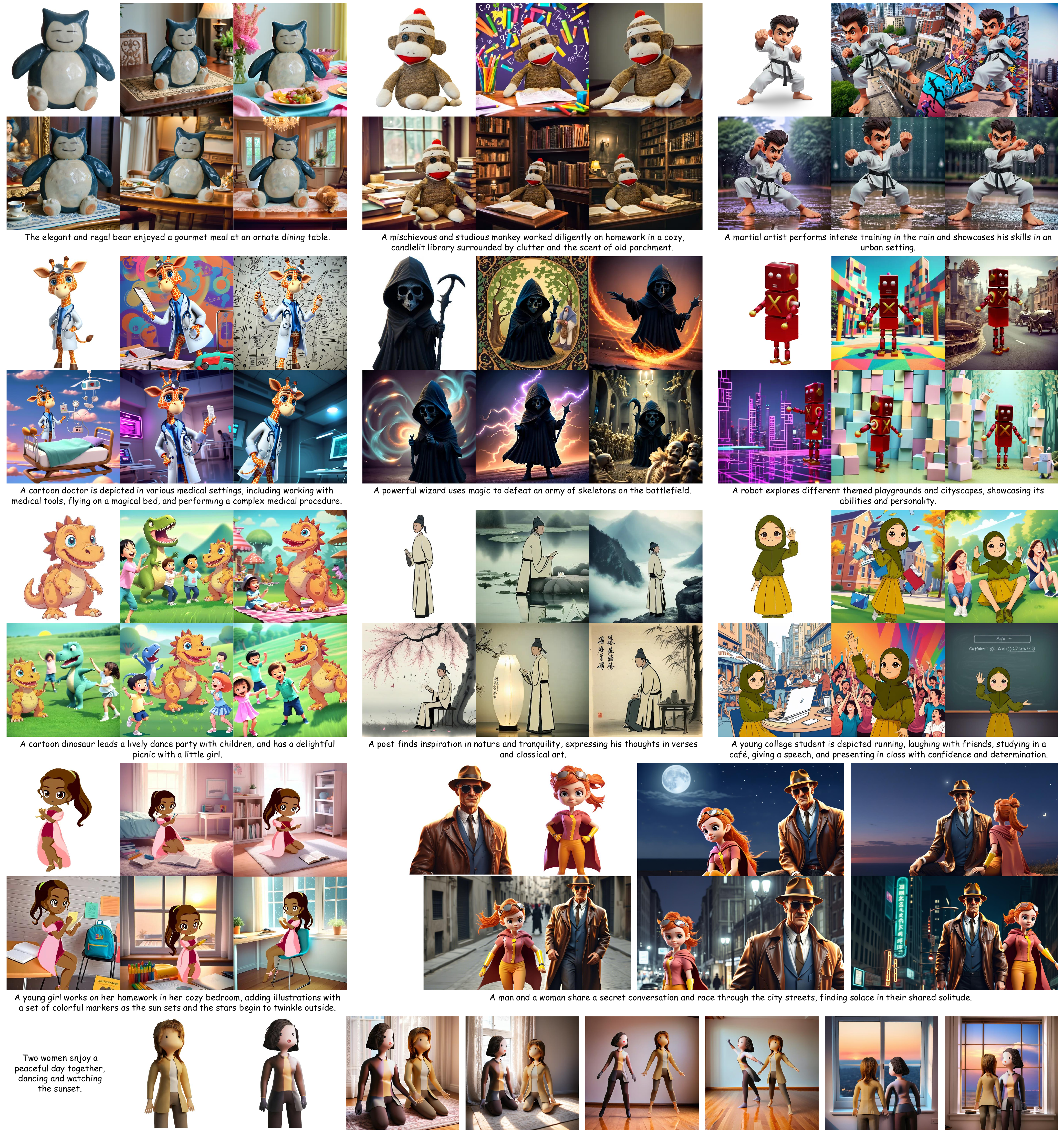}
	\caption{
		\textbf{Example generations \uppercase\expandafter{\romannumeral3} from AnyStory.} 
	}
	\label{fig:example-3}
\end{figure*}

\section{Experiments}

\subsection{Setup}

We use the stable diffusion XL~\cite{podell2023sdxl} as the base model. The CLIP image encoder employed is the OpenAI's \texttt{clip-vit-huge-patch14}. Both the subject QFormer and the routing QFormer consist of 4 layers. The input image resolution for ReferenceNet is $512\times512$. All training is conducted on 8 A100 GPUs with a batch size of 64, utilizing the AdamW~\cite{loshchilov2017decoupled} optimizer with a learning rate of 1e-4.
In order to facilitate classifier-free guidance sampling~\cite{ho2022classifier}, we drop the CLIP subject conditioning during training on 10\% of the images. During the inference process, we employ 25 steps of EDM~\cite{karras2022elucidating} sampling and a 7.5 classifier-free guidance scale, and to achieve more realistic image generation, we employ the \textit{RealVisXL V4.0} model from \textit{huggingface}.

\subsection{Effect of ReferenceNet encoder}

\figref{fig:referencenet-encoder} illustrates the effectiveness of the ReferenceNet encoder, which enhances the preservation of fine details in the subject compared to using only CLIP vision encoder. However, it is also evident that using ReferenceNet alone does not yield satisfactory results. In fact, in our extensive testing, we found that the ReferenceNet encoder only achieves alignment of the subject details and does not guide subject generation. We still need to rely on CLIP-encoded features, which are well-aligned with text embeddings, to trigger subject generation.

\subsection{Effect of the decoupled instance-aware router}

\figref{fig:router-effect} demonstrates the effectiveness of the proposed router, which can effectively avoid feature blending between subjects in multi-subject generation. Additionally, we observe that the use of the router in single-subject settings also improves the quality of generated images, particularly in the image background. This is because the router restricts the influence area of subject conditions, thereby reducing the potential bias inherent in subject features (\eg, simple white background preference learned from a large amount of 3D rendering data) on the quality of generated images.

\figref{fig:router-visualize} visualizes the routing maps of the diffusion model at different time steps during the denoising process. These results demonstrate that the proposed router can accurately perceive and locate the effect regions of each subject condition during the denoising process. The displayed routing maps are similar to image segmentation masks, indicating the potential for achieving guided image segmentation based on reference images through denoising U-Net and trained routers. Additionally, as mentioned in \secref{sec:3.4}, despite our router being trained predominantly on human-centric datasets, it generalizes well to general subjects such as the cartoon dinosaur in \figref{fig:router-visualize}. We attribute this to the powerful generalization capability of the CLIP model and the highly compressed single-token routing features.

\figref{fig:router-coarse-refine}  demonstrates the effectiveness of modeling the router as a miniature image segmentation decoder. Compared to the coarse routing map obtained by a simple dot product, the refined routing map through a lightweight masked cross-attention module can more accurately predict the potential position of each subject.

\subsection{Example generations}
In \figref{fig:example-1}, \figref{fig:example-2}, and \figref{fig:example-3}, we visualize further results of our approach, demonstrating its outstanding performance in preserving subject details, aligning text prompts, and enabling multi-subject personalization.

%% file: sections/concl.tex
\section{Conclusion}

We propose AnyStory, a unified method for personalized generation of both single and multiple subjects. AnyStory utilizes a universal and powerful ReferenceNet in addition to a CLIP vision encoder to achieve high-fidelity subject encoding, and employs a decoupled, instance-aware routing module for flexible and accurate single/multiple subject condition injection. Experimental results demonstrate that our method excels in retaining subject details, aligning with textual descriptions, and personalizing for multiple subjects. 

\noindent\textbf{Limitations and future work.} Currently, AnyStory is unable to generate personalized backgrounds for images. However, maintaining consistency in the image background is equally important in sequential image generation. In the future, we will expand AnyStory's control capabilities from the subject domain to the background domain. Additionally, the copy-paste effect still exists in the subjects generated by AnyStory, and we aim to mitigate this further in the future through data augmentation and the use of more powerful text-to-image generation models.

%% file: sections/appendix.tex
\section{Referenced subject images and URLs}
This section consolidates the sources of the referenced subject images in this paper. We extend our gratitude to the owners of these images for sharing their valuable assets.

\begin{longtable}{m{2cm}<{\centering}|m{15cm}<{\centering}}
	\hline
	Reference & URL  \\
	\hline
	\endfirsthead
	\hline
	Reference & URL \\
	\hline
	\endhead
	\hline
	\endfoot
	\hline
	\endlastfoot
	 ~ & ~\\
\includegraphics[width=0.5\linewidth]{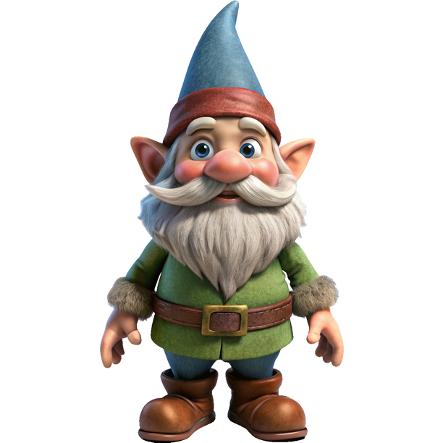} & \small \url{https://pixabay.com/illustrations/ai-generated-dwarf-story-fantasy-8697130/}  \\
\includegraphics[width=0.5\linewidth]{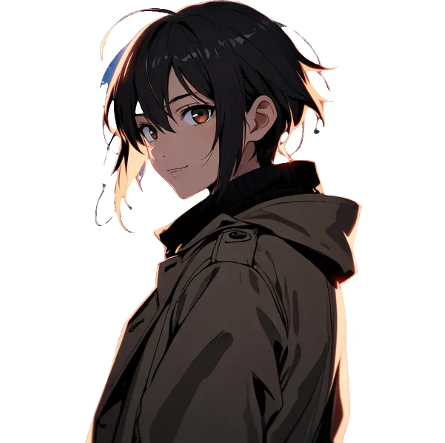} & \small \url{https://pixabay.com/illustrations/girl-coat-night-night-city-8836068/}  \\
\includegraphics[width=0.5\linewidth]{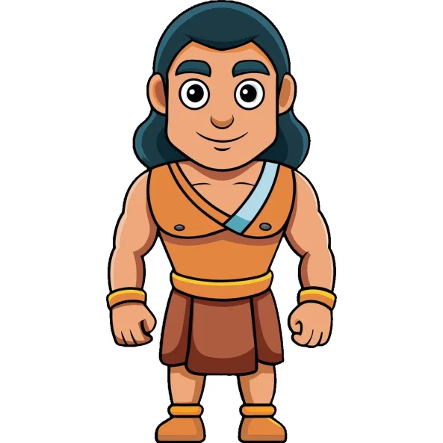} & \small \url{https://pixabay.com/vectors/man-warrior-art-character-cartoon-9093563/}  \\
\includegraphics[width=0.5\linewidth]{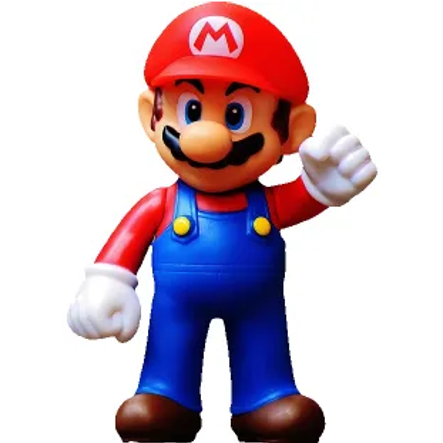} & \small \url{https://pixabay.com/photos/mario-figure-game-nintendo-super-1558068/}  \\
\includegraphics[width=0.5\linewidth]{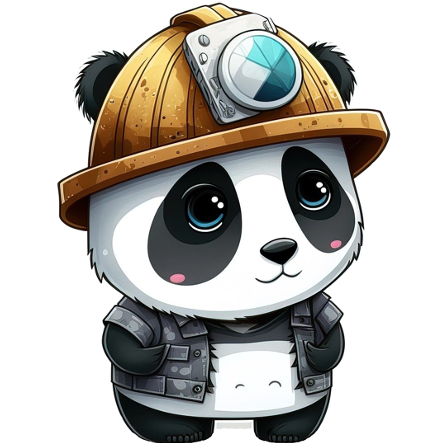} & \small \url{https://pixabay.com/illustrations/panda-cartoon-2d-art-character-7918136/}  \\
\includegraphics[width=0.5\linewidth]{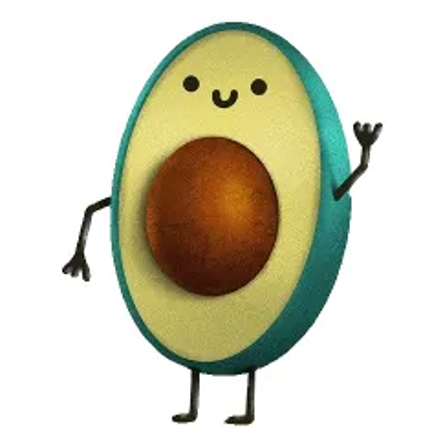} & \small \url{https://pixabay.com/illustrations/avocado-food-fruit-6931344/}  \\
\includegraphics[width=0.5\linewidth]{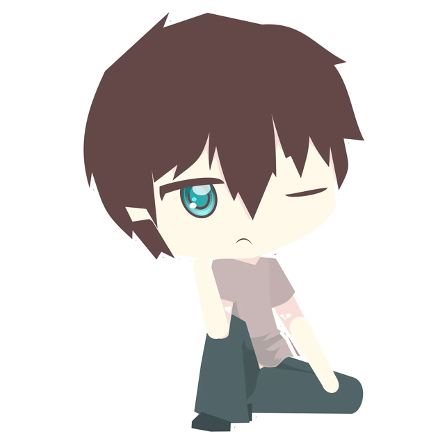} & \small \url{https://pixabay.com/vectors/guy-anime-cartoon-chibi-character-7330732/}  \\
\includegraphics[width=0.5\linewidth]{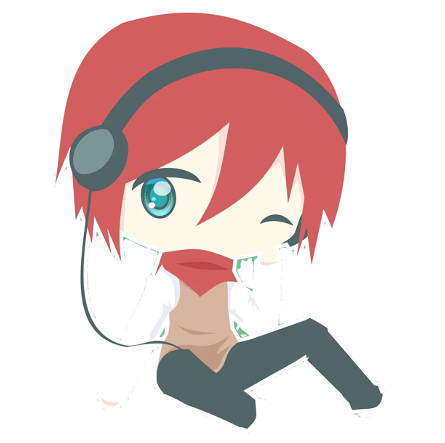} & \small \url{	https://pixabay.com/vectors/guy-anime-cartoon-chibi-character-7330788/}  \\
\includegraphics[width=0.5\linewidth]{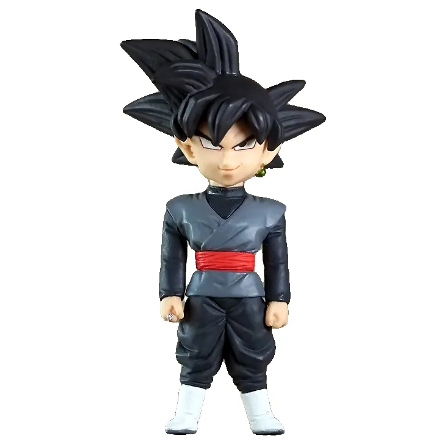} & \small \url{	https://pixabay.com/photos/young-male-man-japanese-anime-3815077/}  \\
\includegraphics[width=0.5\linewidth]{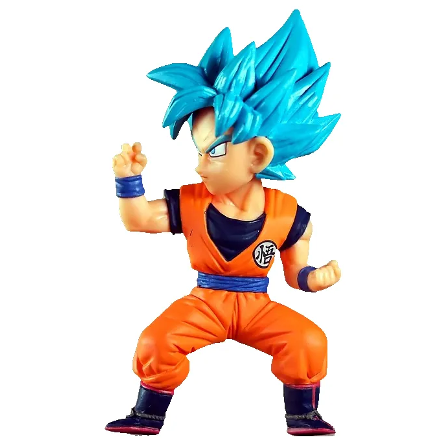} & \small \url{https://pixabay.com/photos/young-male-man-japanese-anime-3816557/}  \\
\includegraphics[width=0.5\linewidth]{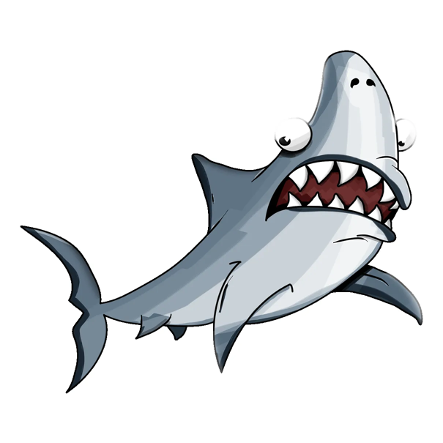} & \small \url{https://pixabay.com/illustrations/shark-jaws-fish-animal-marine-life-2317422/}  \\
\includegraphics[width=0.5\linewidth]{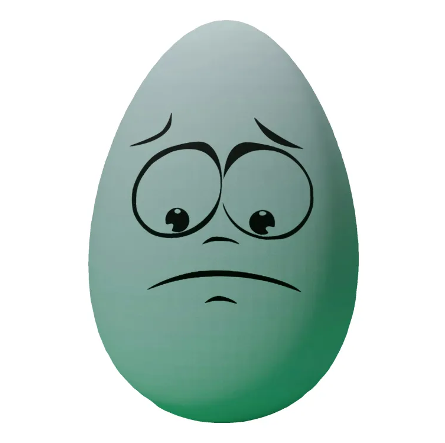} & \small \url{https://unsplash.com/photos/white-egg-with-face-illustration-WtolM5hsj14}  \\
\includegraphics[width=0.5\linewidth]{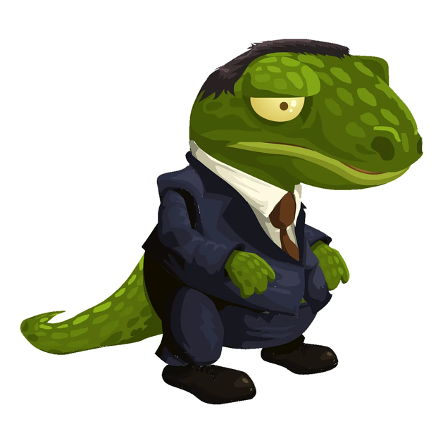} & \small \url{https://pixabay.com/vectors/alligator-crocodile-suit-cartoon-576481/}  \\
\includegraphics[width=0.5\linewidth]{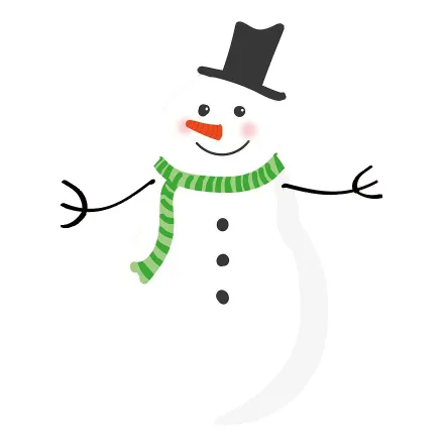} & \small \url{https://pixabay.com/illustrations/snowman-winter-christmas-time-snow-7583640/}  \\
\includegraphics[width=0.5\linewidth]{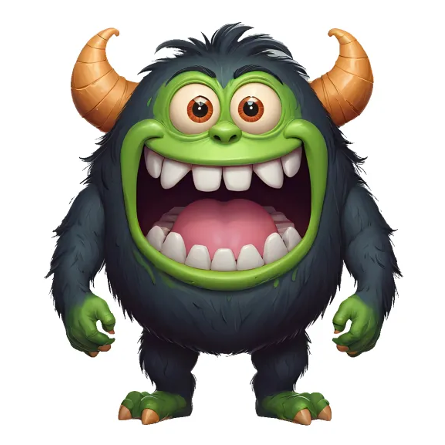} & \small \url{	https://pixabay.com/illustrations/monster-cartoon-funny-creature-8534186/}  \\
\includegraphics[width=0.5\linewidth]{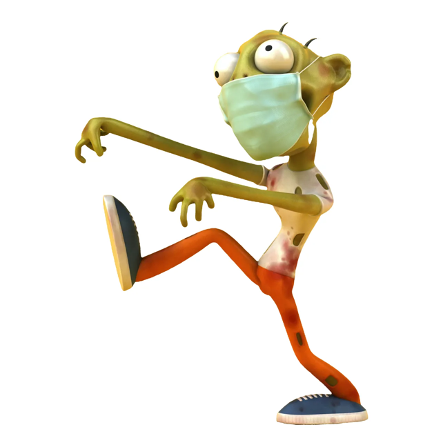} & \small \url{https://unsplash.com/photos/a-cartoon-character-wearing-a-face-mask-and-running-6-adg66qleM}  \\
\includegraphics[width=0.5\linewidth]{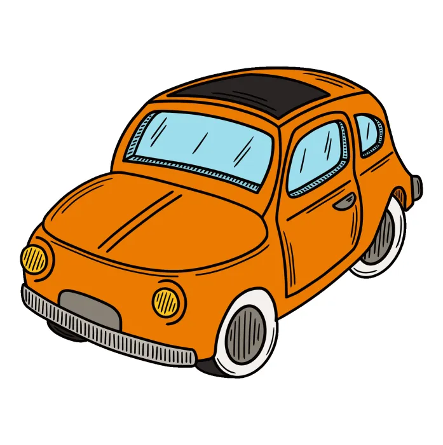} & \small \url{https://pixabay.com/illustrations/car-vehicle-drive-transportation-8316057/}  \\
\includegraphics[width=0.5\linewidth]{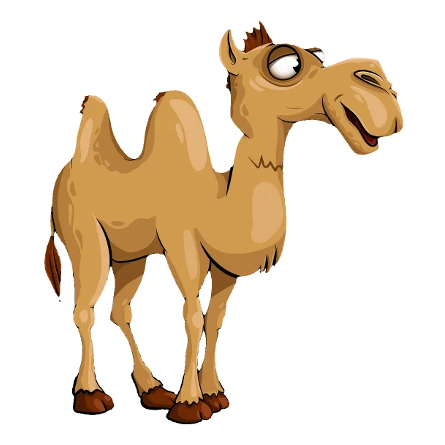} & \small \url{https://pixabay.com/vectors/camel-desert-two-humped-animal-7751098/}  \\
\includegraphics[width=0.5\linewidth]{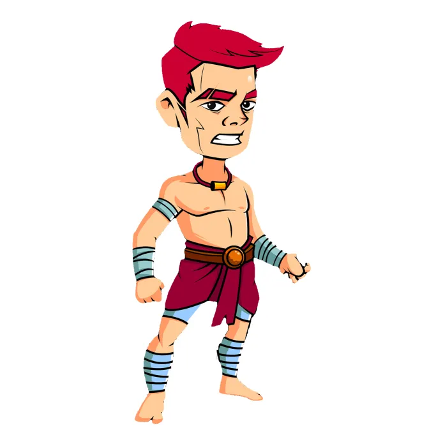} & \small \url{https://pixabay.com/illustrations/cartoon-samurai-characters-4790355/}  \\
\includegraphics[width=0.5\linewidth]{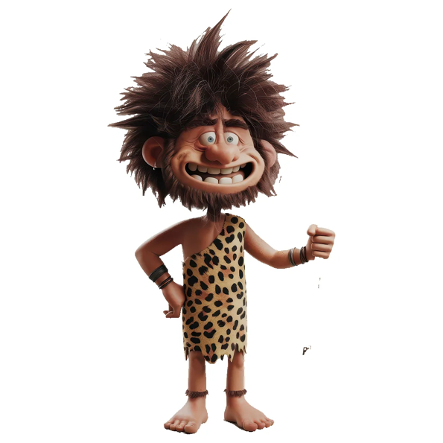} & \small \url{https://pixabay.com/illustrations/caveman-prehistoric-character-9211043/}  \\
\includegraphics[width=0.5\linewidth]{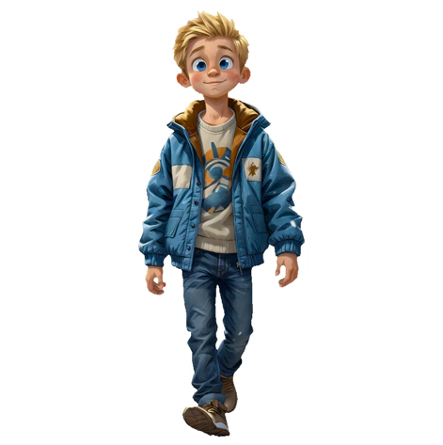} & \small \url{	https://pixabay.com/illustrations/boy-walk-nature-anime-smile-8350034/}  \\
\includegraphics[width=0.5\linewidth]{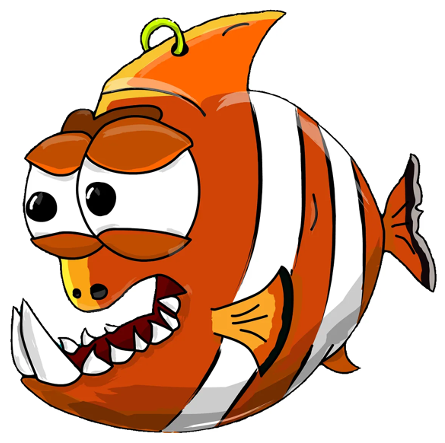} & \small \url{https://pixabay.com/illustrations/fish-jaw-angry-cartoon-parrot-fish-1402423/}  \\
\includegraphics[width=0.5\linewidth]{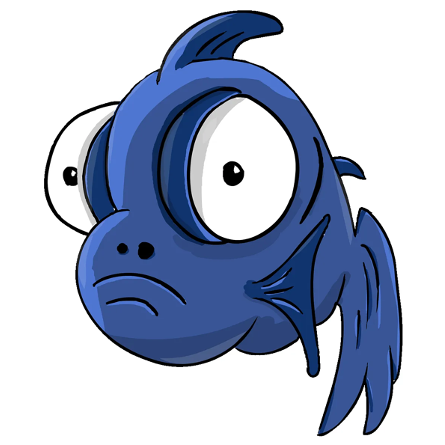} & \small \url{https://pixabay.com/illustrations/fish-telescope-fish-cartoon-1450768/}  \\
\includegraphics[width=0.5\linewidth]{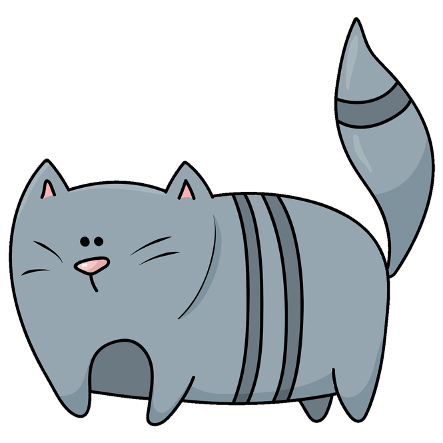} & \small \url{https://pixabay.com/vectors/cat-pet-animal-kitty-kitten-cute-6484941/}  \\
\includegraphics[width=0.5\linewidth]{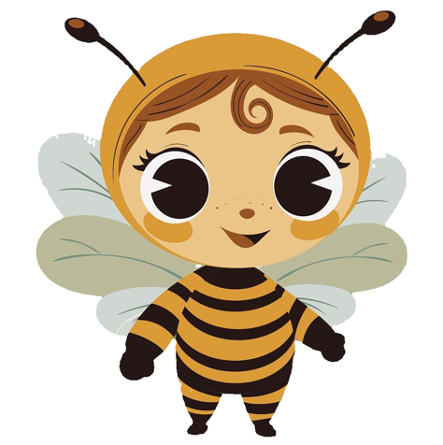} & \small \url{https://pixabay.com/vectors/child-costume-bee-character-8320341/}  \\
\includegraphics[width=0.5\linewidth]{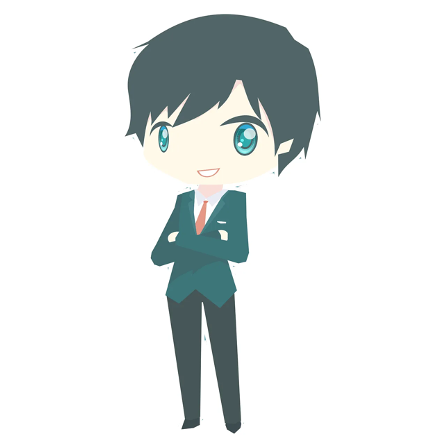} & \small \url{https://pixabay.com/vectors/guy-anime-cartoon-chibi-character-7330758/}  \\
\includegraphics[width=0.5\linewidth]{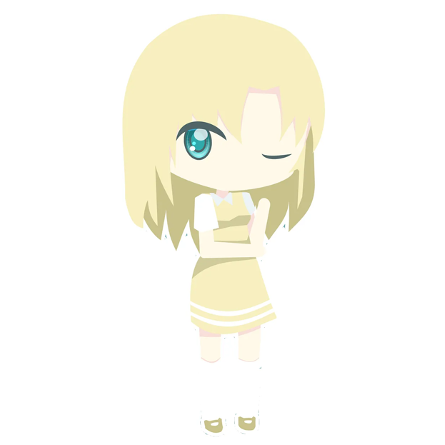} & \small \url{	https://pixabay.com/vectors/girl-anime-chibi-cartoon-character-7346667/}  \\
\includegraphics[width=0.5\linewidth]{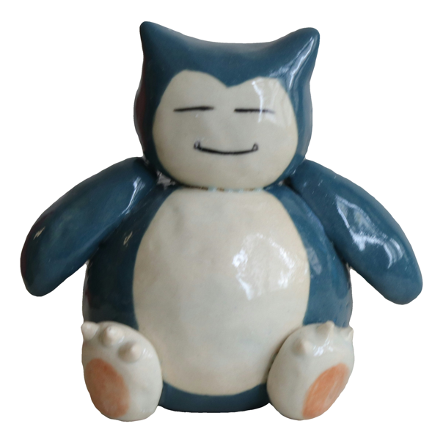} & \small \url{https://unsplash.com/photos/white-and-blue-cat-figurine-u3ZUSIH_eis}  \\
\includegraphics[width=0.5\linewidth]{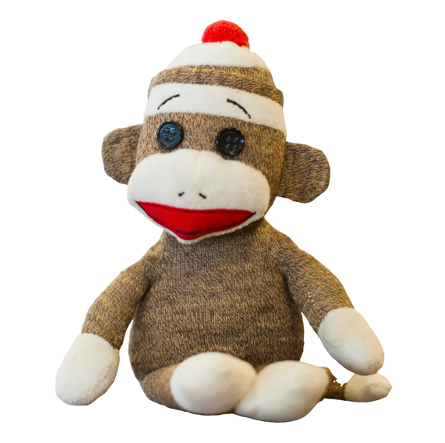} & \small \url{	https://unsplash.com/photos/sock-monkey-plush-toy-on-brown-panel-5INN0oj12u4}  \\
\includegraphics[width=0.5\linewidth]{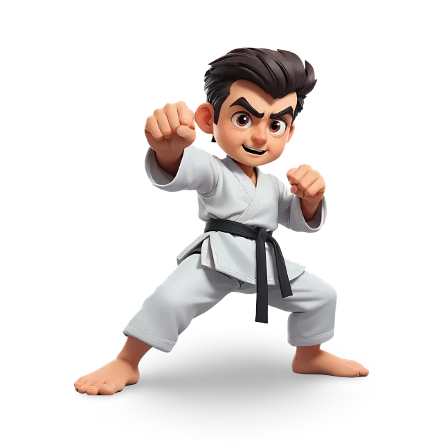} & \small \url{https://pixabay.com/illustrations/karate-fighter-cartoon-character-8537724/}  \\
\includegraphics[width=0.5\linewidth]{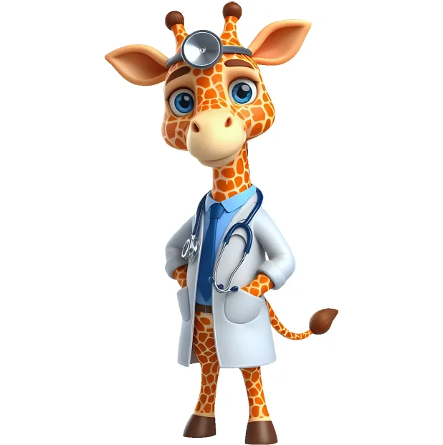} & \small \url{https://pixabay.com/illustrations/ai-generated-giraffe-doctor-8647702/}  \\
\includegraphics[width=0.5\linewidth]{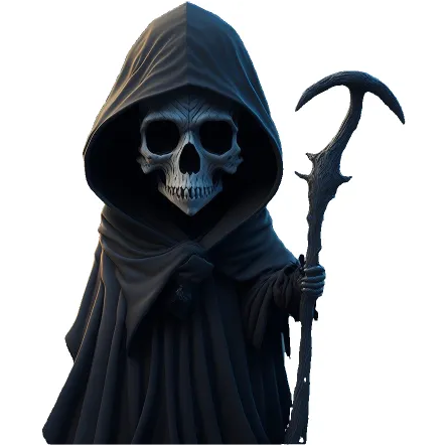} & \small \url{https://pixabay.com/illustrations/ai-generated-skull-character-8124354/}  \\
\includegraphics[width=0.5\linewidth]{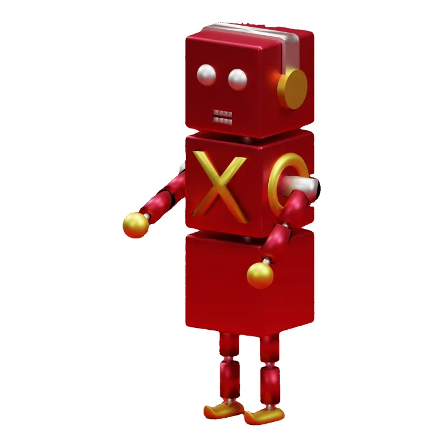} & \small \url{https://unsplash.com/photos/a-red-robot-is-standing-on-a-pink-background-unt3066GV-E}  \\
\includegraphics[width=0.5\linewidth]{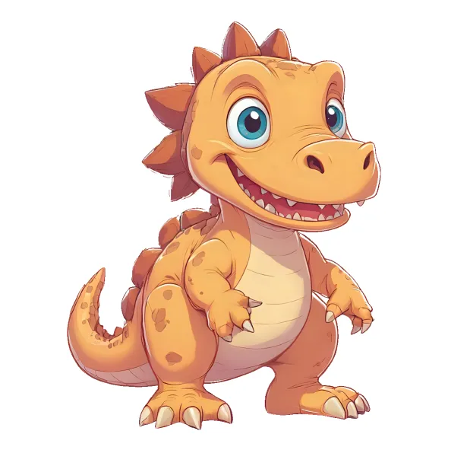} & \small \url{	https://pixabay.com/illustrations/cartoon-dinosaur-dragon-animal-8539364/}  \\
\includegraphics[width=0.5\linewidth]{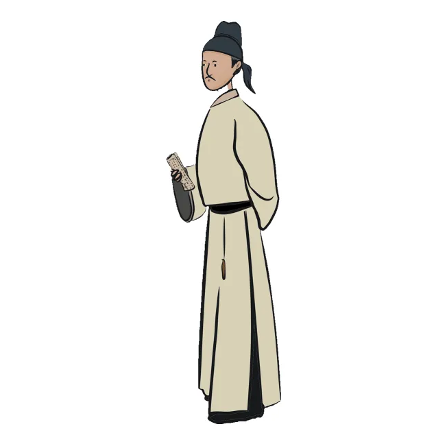} & \small \url{	https://pixabay.com/illustrations/man-book-read-hanfu-chinese-hanfu-7364886/}  \\
\includegraphics[width=0.5\linewidth]{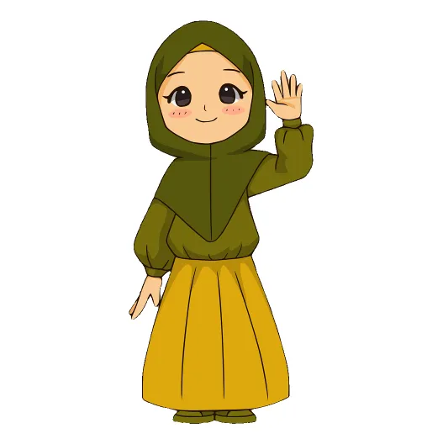} & \small \url{https://pixabay.com/vectors/muslim-hijab-child-cartoon-doodle-7747745/}  \\
\includegraphics[width=0.5\linewidth]{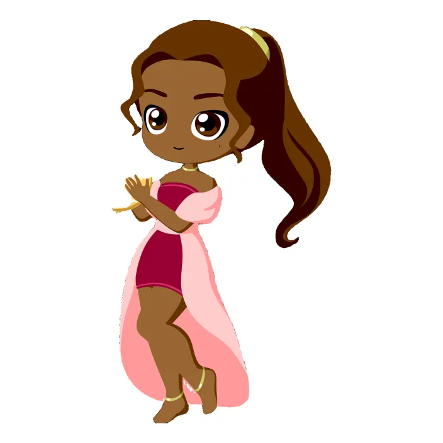} & \small \url{https://pixabay.com/illustrations/tambourine-musician-woman-character-9073083/}  \\
\includegraphics[width=0.5\linewidth]{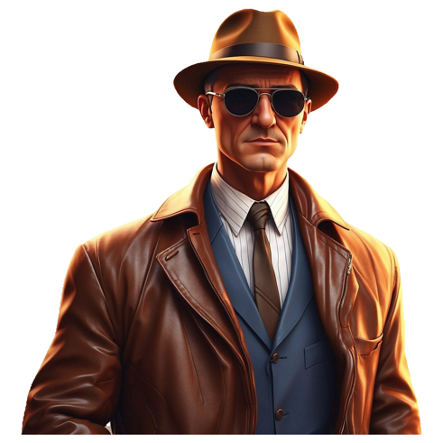} & \small \url{	https://pixabay.com/illustrations/ai-generated-man-agent-character-9050849/}  \\
\includegraphics[width=0.5\linewidth]{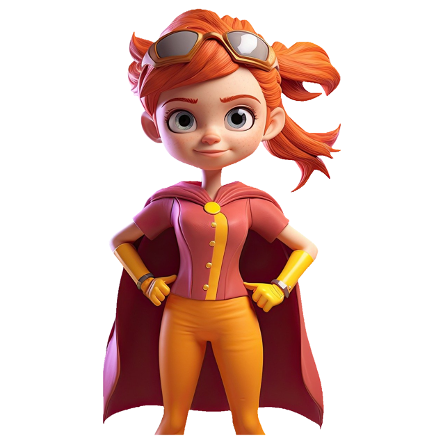} & \small \url{https://pixabay.com/illustrations/ai-generated-superhero-hero-heroine-7977051/}  \\
\includegraphics[width=0.5\linewidth]{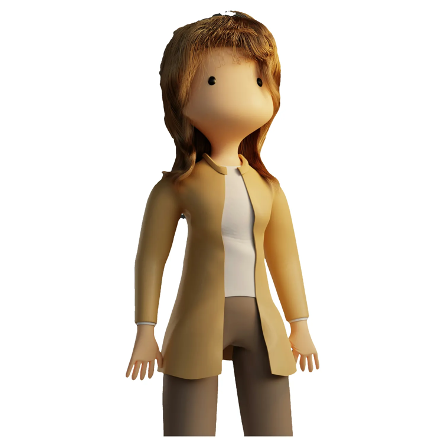} & \small \url{https://unsplash.com/photos/a-woman-in-a-tan-jacket-and-tan-pants-QVyAUDUOlMw}  \\
\includegraphics[width=0.5\linewidth]{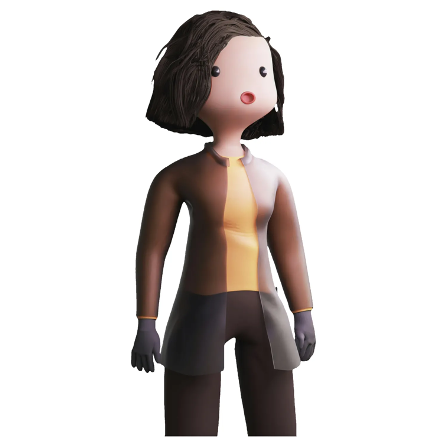} & \small \url{https://unsplash.com/photos/a-woman-in-a-yellow-shirt-and-black-pants-rdHrrFA1KKg}  \\
\includegraphics[width=0.5\linewidth]{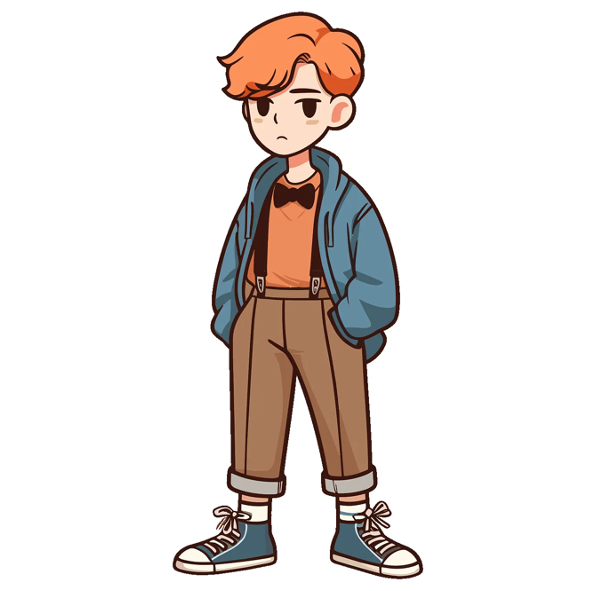} & \small \url{https://pixabay.com/vectors/fashion-boy-cartoon-spring-summer-8515751/}  \\
\includegraphics[width=0.5\linewidth]{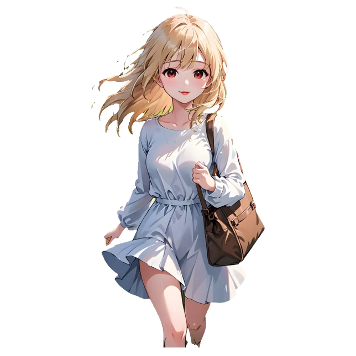} & \small \url{https://pixabay.com/illustrations/woman-girl-fashion-model-female-8859569/}  \\
\includegraphics[width=0.5\linewidth]{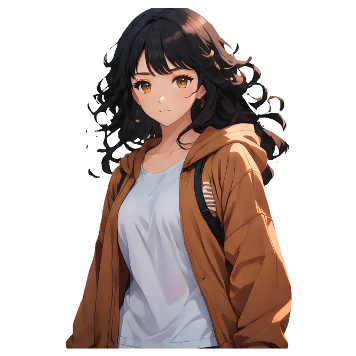} & \small \url{	https://pixabay.com/illustrations/woman-cartoon-character-anime-8926994/}  \\
\includegraphics[width=0.5\linewidth]{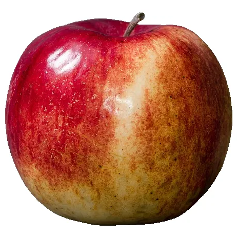} & \small \url{https://pixabay.com/photos/apple-red-delicious-fruit-vitamins-256268/}  \\
\includegraphics[width=0.5\linewidth]{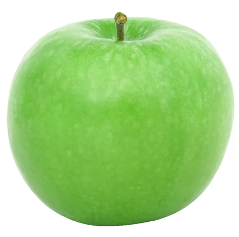} & \small \url{tps://pixabay.com/photos/apple-food-fresh-fruit-green-1239300/}  \\
\includegraphics[width=0.5\linewidth]{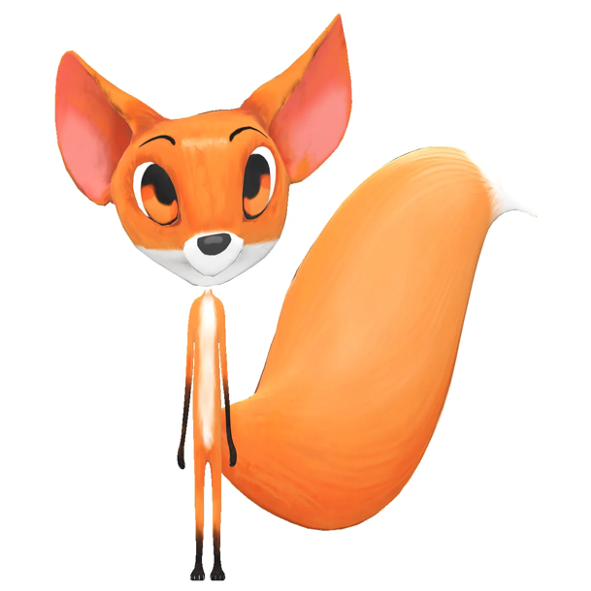} & \small \url{	https://pixabay.com/illustrations/fox-animal-wildlife-wild-mammal-9267914/}  \\
\includegraphics[width=0.5\linewidth]{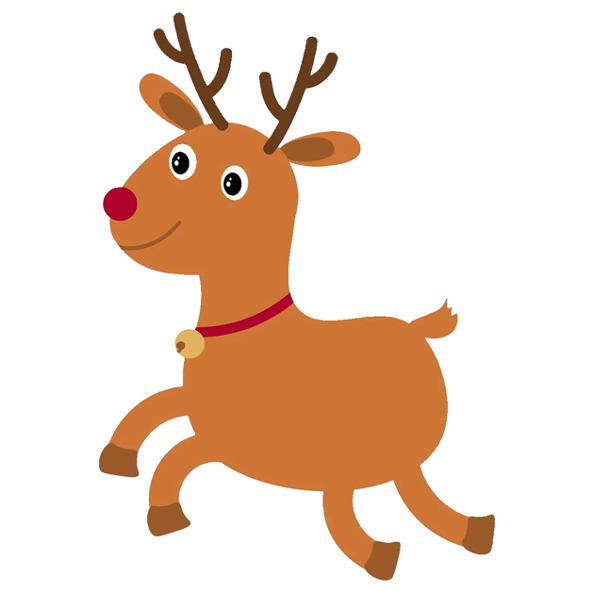} & \small \url{https://pixabay.com/illustrations/christmas-deer-animal-rudolph-8380345/}  \\
\includegraphics[width=0.5\linewidth]{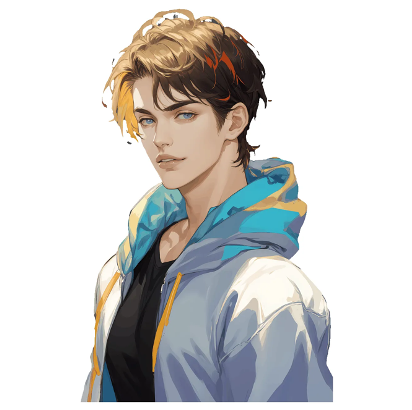} & \small \url{https://pixabay.com/illustrations/ai-generated-man-portrait-7953120/}  \\
\includegraphics[width=0.5\linewidth]{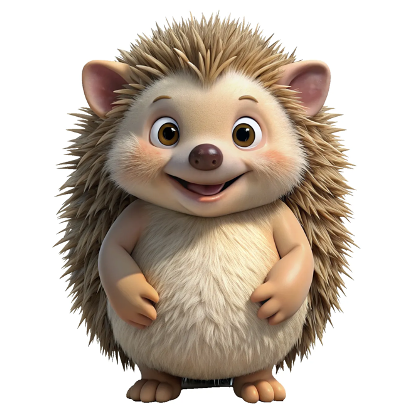} & \small \url{https://pixabay.com/illustrations/created-by-ai-hedgehog-cartoon-8635844/}  \\
\includegraphics[width=0.5\linewidth]{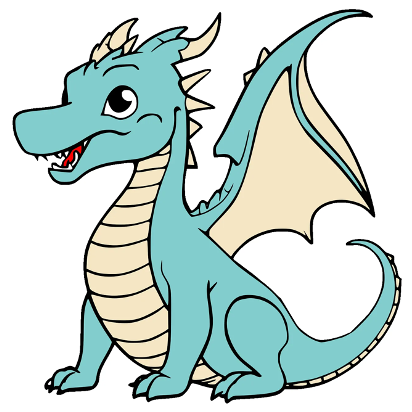} & \small \url{https://pixabay.com/vectors/dragon-creature-baby-dragon-8480029/}  \\
\includegraphics[width=0.5\linewidth]{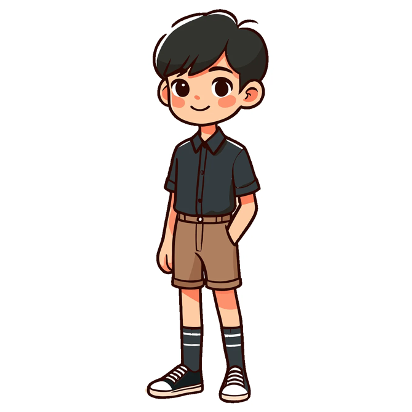} & \small \url{https://pixabay.com/vectors/boy-cartoon-fashion-chibi-kawaii-8515729/}  \\
\includegraphics[width=0.5\linewidth]{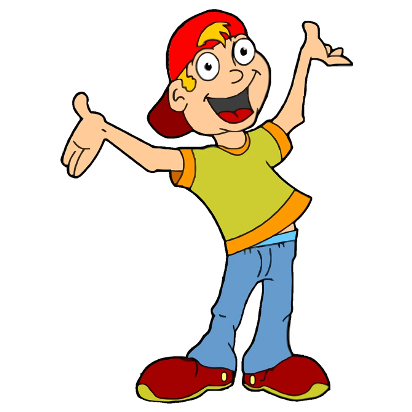} & \small \url{https://pixabay.com/vectors/blonde-boy-cartoon-character-comic-1300066/}  \\
\end{longtable}